\documentclass[10pt,twocolumn,letterpaper]{article}
\usepackage{preprint}
\usepackage[T1]{fontenc}
\usepackage{array,tabularx}
\usepackage{placeins}

\newsavebox{\checkedtablebox}
\newenvironment{checkedtable}[1]{%
  \def\checkedtablelimit{#1}%
  \begin{lrbox}{\checkedtablebox}%
}{%
  \end{lrbox}%
  \ifdim\wd\checkedtablebox>\checkedtablelimit\relax
    \PackageError{hogwam-layout}{Table exceeds its allocated width}%
      {Split the table or shorten its headings to fit the text width.}%
  \fi
  \usebox{\checkedtablebox}%
}

\usepackage{capt-of}
\definecolor{linkblue}{rgb}{0.21,0.49,0.74}
\usepackage[breaklinks,colorlinks,allcolors=linkblue]{hyperref}
\title{HOG-WAM: Improving Visual Robustness through Future Gradient Prediction}
\author{%
Boxuan Dong$^{1}$ \quad Yupeng Qian$^{1}$ \quad Ying Wu$^{1}$ \quad Heshuai Li$^{1}$\\
Junhao Tan$^{1}$ \quad Zhenxiong Tan$^{2}$ \quad Weijia Wu$^{2,*}$ \quad Weihao Yu$^{1,*}$\\
$^{1}$Peking University \qquad $^{2}$National University of Singapore\\[0.4em]
{\small Code: \url{https://github.com/iPPLab/HOG-WAM}.}
}

\hypersetup{pdftitle={HOG-WAM: Improving Visual Robustness through Future Gradient Prediction},
pdfauthor={Boxuan Dong, Yupeng Qian, Ying Wu, Heshuai Li, Junhao Tan, Zhenxiong Tan, Weijia Wu, Weihao Yu}}
\date{}
\begin{document}
\maketitle
\begingroup
\renewcommand{\thefootnote}{\fnsymbol{footnote}}
\footnotetext[1]{Corresponding authors.}
\endgroup

\begin{abstract}
World-action models jointly predict future observations and actions, yet visual corruption can disrupt the resulting policies.
We introduce HOG-WAM, which adds future Histograms of Oriented Gradients (HOG) as structural supervision derived directly from demonstration images.
An auxiliary HOG diffusion transformer (DiT) module reads RGB features through asymmetric attention, allowing its objective to shape the visual representation used for control.
The module is used only during training and removed at inference, preserving FastWAM's RGB-to-action architecture.
We evaluate four controlled observation corruptions through a visual-corruption extension of RoboTwin, termed RoboTwin-Plus; HOG-WAM achieves the highest success among the compared policies under all four.
Mixed-noise success increases from 78.86\% to 82.86\% relative to FastWAM, with comparable performance on clean scenes.
On LIBERO Plus, overall success increases from 74.49\% to 75.26\%, with the largest gain under sensor noise, from 74.14\% to 78.89\%.
Across four Franka tasks, average success increases from 40\% to 59\% under clean observations and from 17\% to 24\% under noisy observations.
A representation ablation favors combining RGB observations with auxiliary HOG prediction over replacing RGB with HOG.
These findings support future-gradient prediction as complementary supervision for visually robust control, with larger absolute gains on physical tasks than in simulation.
\end{abstract}

\begin{figure*}[t]
    \centering
    \includegraphics[width=\textwidth]{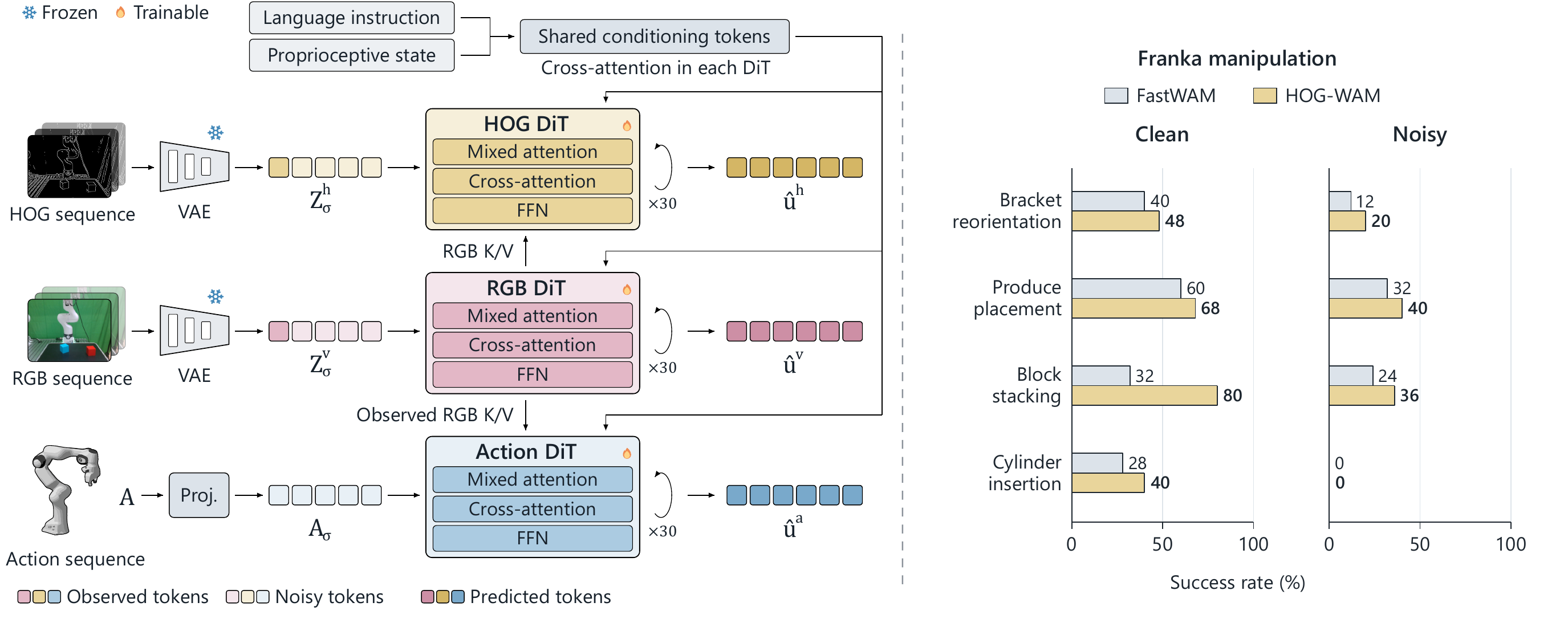}
    \caption{HOG-WAM architecture and real-robot performance. \textbf{Left.} An auxiliary HOG DiT provides future-gradient supervision to the RGB backbone through asymmetric mixed attention during training. Language and proprioception condition all three DiT modules through cross-attention. Inference uses the RGB and action modules. Snowflakes and flames mark frozen and trainable components, and $\times30$ denotes transformer depth. VAE denotes variational autoencoder, FFN feed-forward network, and Proj. projection. \textbf{Right.} FastWAM and HOG-WAM success rates on four Franka manipulation tasks under clean and noisy observations, with 25 trials per task, model, and condition. Noise follows the RoboTwin-Plus mixed-noise operator.}
    \label{fig:overview}
\end{figure*}

\section{Introduction}
\label{sec:introduction}
Robotic manipulation requires recognizing task-relevant objects and anticipating how their spatial relations change during interaction.
Vision-language-action policies use pretrained visual and language representations to map observations to actions~\cite{brohan2023rt2,kim2024openvla,black2024pi0}.
World-action models (WAMs) additionally learn to predict future observations, making demonstrated scene evolution a source of policy supervision~\cite{ye2026dreamzero,li2026lingbot,bi2025motus}.
Fast-WAM~\cite{yuan2026fastwam}, denoted FastWAM here, preserves this predictive objective during training while generating actions from the current observation at inference.
The prediction target therefore determines which visual information this training objective emphasizes.

RGB prediction supervises object structure together with color, texture, and illumination.
These appearance cues help identify objects, but they also vary with lighting and sensor conditions while the manipulation goal remains unchanged.
An RGB objective alone thus ties the learned features to appearance factors that can change at deployment.
For placement, stacking, and insertion, local boundaries describe the image-space arrangement of objects and their surroundings.
Requiring a visual representation to predict these boundaries provides an explicit structural learning signal alongside RGB reconstruction.
The question is how to introduce this signal while retaining the appearance information needed for language-conditioned control.

\paragraph{Why gradients?}
Histograms of Oriented Gradients (HOG)~\cite{dalal2005hog} summarize local edge orientations through spatial pooling and contrast normalization.
MaskFeat~\cite{wei2022maskfeat} demonstrates that predicting HOG features in masked image and video regions can learn useful visual representations, identifying local contrast normalization as an important component.
This motivates a different prediction problem for manipulation: learning the gradient structure of future demonstration frames alongside actions.
The target emphasizes where local intensity transitions occur and how they are oriented, making spatial organization explicit without object annotations or a learned target generator.

\paragraph{Why training-only supervision?}
A descriptor useful for supervising a representation need not replace its sensory input.
Our HOG raster compresses orientation statistics and removes chromatic information; RGB retains cues for object identity and instruction grounding.
WAM4D~\cite{li2026wam4d} offers a related architectural principle: future-depth supervision shapes the visual features used for control through a removable readout.
We apply this principle to deterministic gradient targets, which require no additional sensor or pretrained estimator.
In \textbf{HOG-WAM}, an auxiliary HOG diffusion transformer (DiT) reads RGB features through asymmetric attention, and its loss updates the RGB backbone.
Action prediction remains independent of HOG tokens, allowing the auxiliary module to be removed at inference while retaining its effect on the learned RGB representation.

We evaluate visual robustness on RoboTwin 2.0~\cite{chen2025robotwin}, LIBERO Plus~\cite{fei2025liberoplus}, and four Franka tasks.
RoboTwin-Plus denotes our visual-corruption extension of the existing RoboTwin evaluation, with fixed tasks and controlled changes to image inputs.
On LIBERO Plus, gains concentrate under sensor noise.
A LIBERO Plus ablation compares HOG as auxiliary supervision with HOG as a replacement for RGB.
The physical experiments show larger absolute gains over FastWAM than the simulation averages, extending the evidence to manipulation with real camera observations.
Our contributions are:
\begin{itemize}
\item A deterministic future-gradient target that makes local structure an explicit supervision signal for world-action learning.
\item An asymmetric coupling of RGB, HOG, and action DiT modules that transfers this supervision to the visual backbone while preserving RGB-based inference.
\item Evidence from controlled visual-corruption tests and physical manipulation, together with a representation ablation that supports complementary HOG supervision over RGB replacement.
\end{itemize}

\section{Related work}
\label{sec:related_work}
\paragraph{Generative models and robot policies.}
Transformers~\cite{vaswani2017attention,dosovitskiy2021vit} provide token-based architectures for visual learning, while diffusion models~\cite{ho2020ddpm}, latent diffusion~\cite{rombach2022ldm}, and DiTs~\cite{peebles2023dit} support conditional image generation.
Flow matching~\cite{lipman2023flowmatching} and rectified flow~\cite{liu2023rectifiedflow} learn velocity fields that transport noise toward data.
Diffusion Policy~\cite{chi2023diffusionpolicy} applies conditional denoising to action sequences.
RT-1 and RT-2~\cite{brohan2023rt1,brohan2023rt2} scale observation-to-action learning, while Octo and OpenVLA~\cite{octo2024,kim2024openvla} provide generalist policies adaptable to new robot settings.
The $\pi_0$ family~\cite{black2024pi0,pi052025} combines pretrained vision-language representations with continuous action generation and heterogeneous training data.
HOG-WAM instead studies how the visual prediction objective within a WAM can improve the representation used for control.

\paragraph{Predictive learning for control.}
UniPi~\cite{du2023unipi} derives actions from text-conditioned video plans, and GR-1~\cite{wu2023gr1} jointly predicts future images and actions after video pretraining.
Video foundation models such as Wan~\cite{wan2025wan} supply visual dynamics priors for recent WAMs.
DreamZero~\cite{ye2026dreamzero}, LingBot-VA~\cite{li2026lingbot}, and MOTUS~\cite{bi2025motus} couple visual dynamics and action generation through joint models.
FastWAM~\cite{yuan2026fastwam} separates future-video training from action inference.
Our auxiliary target builds on this separation, changing what the RGB representation is trained to predict.

\paragraph{Structural supervision in world models.}
TesserAct~\cite{zhen2025tesseract} models RGB, depth, and normals, while Kinema4D~\cite{xu2026kinema4d} uses kinematic robot trajectories to condition RGB and point-map simulation.
WAM4D~\cite{li2026wam4d} transfers geometric priors through future-depth readouts that are removed at inference.
MaskWAM~\cite{yu2026maskwam} uses mask prompting and prediction, FlowWAM~\cite{chen2026flowwam} adopts optical flow as an action representation, and Robust-WAM~\cite{yan2026robustwam} aligns action-stream features with semantic targets.
DreamWAM~\cite{yuan2026dreamwam} combines appearance, motion, geometry, and semantics.
HOG-WAM complements these learned or task-specific targets with local gradient statistics computed directly from each demonstration frame.

\paragraph{Masked prediction and gradient targets.}
The target representation determines the information emphasized by a predictive objective.
BEiT~\cite{bao2022beit} predicts discrete visual tokens; MAE and SimMIM~\cite{he2022mae,xie2022simmim} reconstruct pixels; VideoMAE~\cite{tong2022videomae} extends masked reconstruction to videos; and I-JEPA~\cite{assran2023ijepa} predicts learned target embeddings.
MaskFeat~\cite{wei2022maskfeat} identifies HOG~\cite{dalal2005hog} as an effective feature-regression target, connecting local shape information with representation learning.
HOG-WAM carries this target-design insight into predictive robot learning: a HOG-derived raster is predicted in the video latent space alongside future RGB and actions.
The structural predictor is an auxiliary training objective, while the policy retains RGB observations for control.

\section{HOG-WAM}
\label{sec:method}
\subsection{Predictive policy learning}
\label{sec:formulation}
At each decision, the policy receives a multi-camera RGB observation $I_0$, an instruction $\ell$, and proprioceptive state $s_0$, and predicts an action chunk
\begin{equation}
A=(a_0,\ldots,a_{K-1})\sim\pi_\theta(\cdot\mid I_0,\ell,s_0).
\end{equation}
Here, $K$ is the action horizon, $a_k\in\mathbb R^{d_a}$ is an action of dimension $d_a$, and $\theta$ denotes trainable policy parameters.
Demonstrations additionally provide $T$ sampled future observations $I_{1:T}$ for predictive training.
FastWAM, on which we build, combines a Wan2.2-TI2V-5B visual DiT module with an action DiT module.
A frozen causal variational autoencoder (VAE) $\mathcal E$ maps the observation sequence to RGB latents $Z^v=\mathcal E(I_{0:T})$.
Language embeddings and a trainable proprioceptive projection condition both streams.
The model jointly learns RGB and action velocity fields, with action queries attending to the observed RGB frame.

HOG-WAM adds a future-gradient objective through an auxiliary DiT that reads RGB features (\cref{fig:overview}).
This makes the structural loss act on the visual representation used for action prediction.

\subsection{Future-gradient targets}
\label{sec:hog_targets}

\begin{figure*}[t]
    \centering
    \includegraphics[width=.70\textwidth]{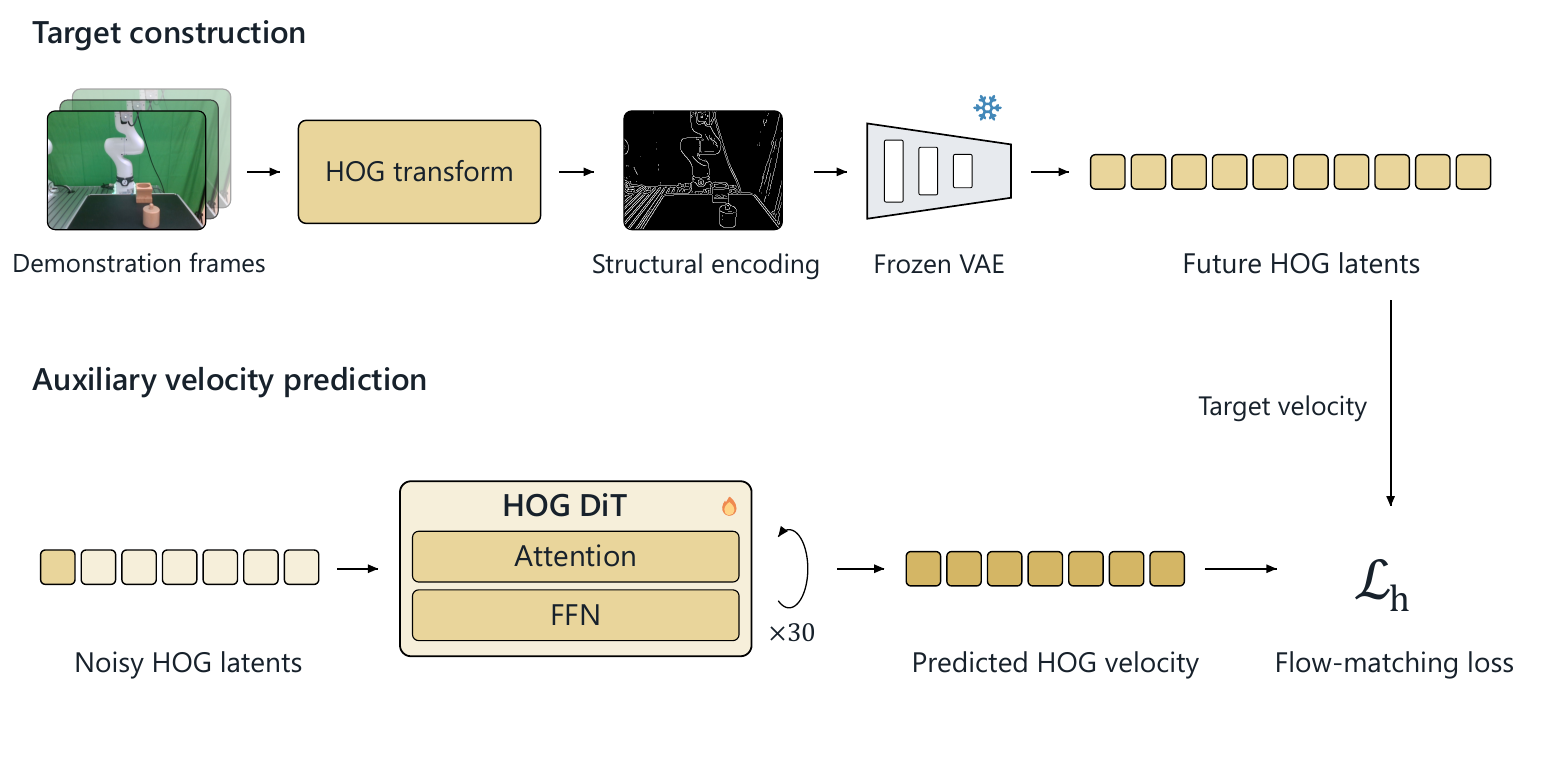}
    \caption{Construction of future-gradient supervision. Demonstration frames provide deterministic structural targets, encoded by the same frozen VAE as RGB. The HOG DiT predicts velocities over noisy HOG latents, and future frames contribute to the auxiliary loss. Monochrome gradients illustrate local boundaries; the three-channel training raster follows \cref{eq:hog_moments}.}
    \label{fig:hog_targets}
\end{figure*}
For each demonstration frame, we compute a deterministic HOG-derived raster $\mathcal H(I_t)$, as illustrated in \cref{fig:hog_targets}.
\paragraph{Local orientation histograms.}
We convert each frame to grayscale and compute horizontal and vertical finite differences.
Gradient magnitudes contribute to nine unsigned orientation bins within $8\times8$ cells, with linear interpolation between neighboring bins.
Overlapping $2\times2$-cell blocks undergo L2-Hys normalization, including clipping at $0.2$ and renormalization.
Averaging contributions from overlapping blocks gives a normalized histogram $h_{qk}$ for each cell $q$ and orientation bin $k$.
This local normalization reduces sensitivity to intensity scale, while the spatial grid retains the arrangement of boundaries.

\paragraph{Rasterization for latent prediction.}
We summarize each normalized histogram by its energy and second angular moments.
For bins $k\in\{0,\ldots,8\}$ with centers $\beta_k=(k+\tfrac12)\pi/9$, the three quantities are
\begin{equation}
\begin{aligned}
e_q&=\sum_k h_{qk},\\
c_q&=\sum_k h_{qk}\cos(2\beta_k),\\
d_q&=\sum_k h_{qk}\sin(2\beta_k).
\end{aligned}
\label{eq:hog_moments}
\end{equation}
The doubled angle represents unsigned orientations consistently, treating directions separated by $\pi$ as equivalent.
The energy channel measures aggregate histogram magnitude; the angular channels encode orientation concentration and direction.
With a small stabilizer $\varepsilon>0$, we set $b=\max(\max_q e_q,\varepsilon)$ and form the cell raster $(2e_q/b-1,c_q/b,d_q/b)$.
We clip its values to $[-1,1]$ and repeat each cell value over its pixels to recover the original image resolution.
This three-channel representation makes gradient supervision compatible with the pretrained video encoder.
The same frozen VAE encodes the resulting sequence:
\begin{equation}
Z^h=\mathcal E\bigl(\mathcal H(I_{0:T})\bigr).
\end{equation}
RGB and HOG targets share the frozen encoder, latent convention, and temporal positions, pairing appearance and structure from the same future demonstration frames.

\paragraph{Role of the gradient target.}
Let $Y$ denote grayscale intensity and $Y'=aY+\delta$ a photometric change with $a>0$ and no clipping.
Finite differences satisfy $\nabla Y'=a\nabla Y$: the offset disappears, orientations are preserved, and local normalization reduces the shared magnitude scale.
HOG therefore directs supervision toward boundary arrangement rather than absolute intensity.
Cellwise orientation pooling further summarizes local structure without requiring exact pixel reconstruction~\cite{dalal2005hog,wei2022maskfeat}.
Additive noise perturbs gradients directly, so we compute targets from clean demonstrations and evaluate the RGB-input policy under corruption.

Future targets connect this structural emphasis to scene evolution.
Their spatial layout changes as objects move, contact one another, or become occluded, so predicting future HOG encourages visual features informative about these changes.
RGB prediction simultaneously preserves appearance detail.
The two objectives assign complementary descriptions to the same observed transition.

\subsection{Asymmetric attention}
\label{sec:attention}

\begin{figure*}[t]
    \centering
    \includegraphics[width=.72\textwidth]{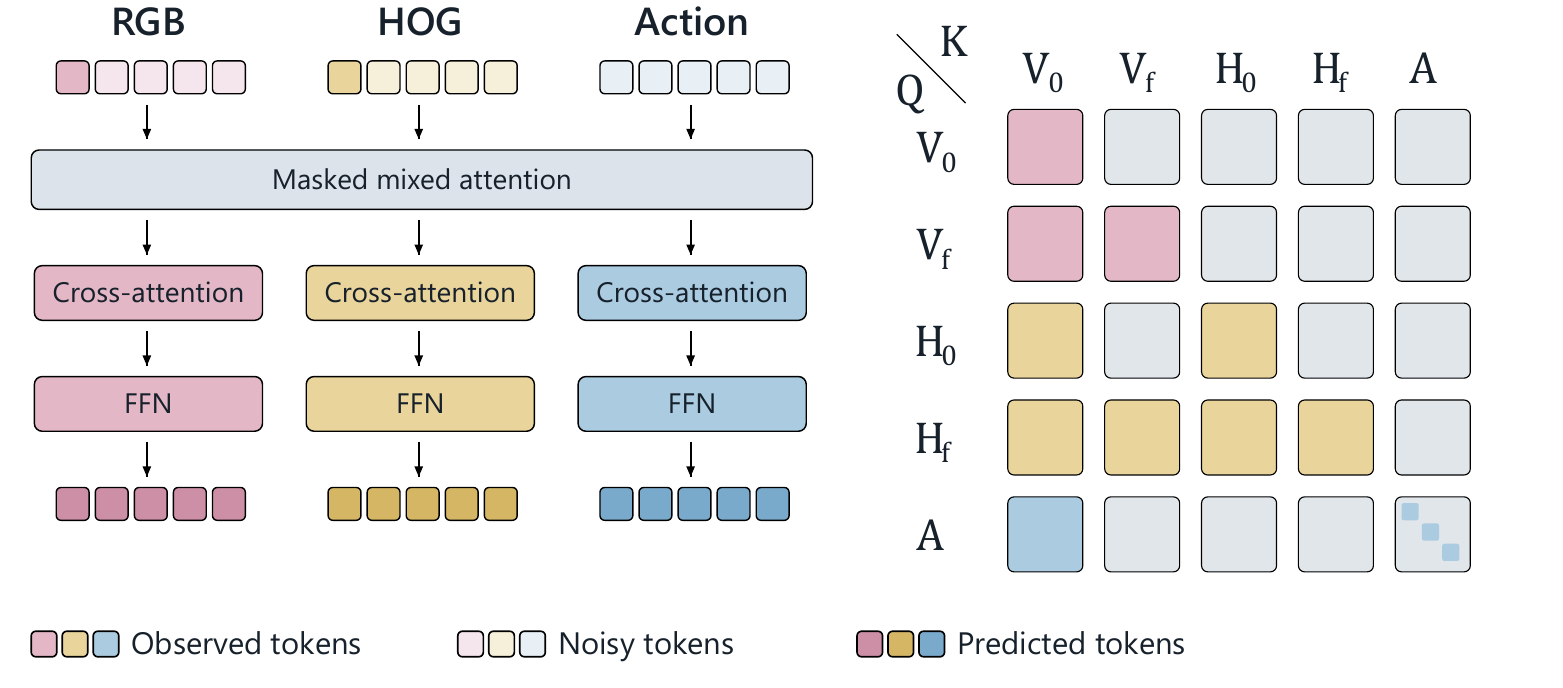}
    \caption{Coupled DiT block and asymmetric attention. Asymmetric mixed attention is followed by contextual cross-attention and modality-specific feed-forward layers. The legend distinguishes observed, noisy, and predicted tokens. In the attention map, $Q$ indexes rows and $K$ indexes columns; $V$ and $H$ denote RGB and HOG, and subscripts $0$ and $f$ identify observed and future frames. Colored cells permit attention; gray cells suppress it. The action diagonal separates independent action copies. FFN denotes feed-forward network; K/V denotes keys and values. The block repeats across 30 layers.}
    \label{fig:attention}
\end{figure*}

\paragraph{Coupled predictors.}
During training, the model contains RGB, HOG, and action DiT modules.
The HOG DiT shares the pretrained initialization of the RGB DiT and has independently optimized parameters.
Each module embeds its noisy state and flow time, with asymmetric mixed attention coupling the streams at every layer.
The RGB and HOG heads predict velocities in the video latent space, and the action head predicts velocities in normalized action coordinates.
Visual latents are divided into $1\times2\times2$ spatiotemporal patches.
The visual and action DiTs use different hidden widths but project their queries, keys, and values to a common attention-head geometry.
Separate feature transformations and output heads preserve each modality's prediction objective.

\paragraph{Attention mask.}
The attention structure connects HOG prediction to RGB learning while preserving the information available to the action policy at inference.
\Cref{fig:attention} shows the permitted query--key connections among RGB, HOG, and action tokens.
RGB queries attend to RGB keys, while HOG queries attend to both RGB and HOG keys.
These visual connections follow a \emph{first-frame-causal} rule: observed-frame queries use observed-frame keys, whereas future queries can use the observed frame and all noisy future frames.
Action queries attend to the observed RGB frame and their own action chunk, with independently sampled action copies kept separate.
All three DiT modules additionally receive language and proprioceptive context through cross-attention.

\paragraph{Why the auxiliary module can be removed.}
Observed RGB tokens use observed RGB keys and shared conditioning at every layer, so they remain independent of future visual and HOG tokens throughout the network.
HOG queries read the same observed-frame RGB tokens as action queries, allowing the HOG loss to update features used for control through the RGB keys and values.
For fixed parameters, these dependencies give
\begin{equation}
\frac{\partial\widehat u^a}{\partial Z^h}=0,
\qquad \nabla_{\theta_v}\mathcal L_h\not\equiv0,
\end{equation}
where $\widehat u^a$ is the predicted action velocity, $\theta_v$ denotes RGB DiT parameters, and $\mathcal L_h$ is the HOG prediction loss.
The policy consequently learns from gradient supervision without depending on a predicted HOG raster during execution.
RGB observations preserve object-identity cues, and auxiliary predictions are excluded from action inputs.
The representation ablation examines the alternative of supplying HOG as the policy input.

\subsection{Joint training and inference}
\label{sec:training}
We train all three DiT modules with flow matching~\cite{lipman2023flowmatching}.
For modality $m\in\{v,h,a\}$, let $X^v=Z^v$, $X^h=Z^h$, and $X^a=A$ denote the clean states.
Gaussian noise $\epsilon^m$ and noise levels $\sigma_m$ are sampled independently for each modality:
\begin{equation}
\begin{aligned}
\widetilde X^m&=(1-\sigma_m)X^m+\sigma_m\epsilon^m,\\
u^m&=\epsilon^m-X^m.
\end{aligned}
\end{equation}
The observed RGB and HOG latent frames remain clean conditioning inputs with flow time zero.
The joint objective supervises future visual frames and valid action positions:
\begin{equation}
\mathcal L=\sum_{m\in\{v,h,a\}}\lambda_m
\,\mathbb E\!\left[w(\sigma_m)\|\widehat u^m-u^m\|_{\mathrm{valid}}^2\right].
\end{equation}
Here, $\|\cdot\|_{\mathrm{valid}}^2$ averages squared error over valid target positions and feature dimensions, and $w$ is the baseline time-dependent weighting function.
We use equal loss weights, $\lambda_v=\lambda_h=\lambda_a=1$.
The RGB, HOG, and action DiT modules and the proprioceptive projection are optimized jointly; the VAE and text encoder remain frozen.
Each visual example is paired with eight copies of its action chunk, with independently sampled noise and flow times.
The action loss is averaged over these copies, as in FastWAM.

At inference, the HOG DiT module and future visual tokens are omitted.
The policy encodes the current RGB observation and conditions action generation on these visual features.
Observed RGB keys and values are computed once per observation and reused across action denoising steps.
For $N$ integration steps with $1=\sigma_N>\cdots>\sigma_0=0$, let $A_i$ be the action chunk at noise level $\sigma_i$. An update takes the form
\begin{equation}
\begin{aligned}
A_{i-1}=A_i+{}& (\sigma_{i-1}-\sigma_i)\\
&\widehat u^a_\theta(A_i,\sigma_i\mid I_0,\ell,s_0).
\end{aligned}
\label{eq:action_integration}
\end{equation}
Starting from Gaussian $A_N$, integration yields an action chunk in normalized coordinates.
The policy converts the chunk to task action units, executes its initial segment, and replans from a new observation.

\FloatBarrier
\section{Experiments}
\label{sec:experiments}

We test whether future-gradient supervision improves control under visual corruption and whether it is more effective as an auxiliary target than as an RGB replacement.
Simulation benchmarks assess controlled visual and task variations; four Franka tasks assess physical manipulation.

\subsection{Experimental setup}
\label{sec:eval_conventions}
FastWAM provides the RGB--action baseline; HOG-WAM adds an auxiliary HOG DiT module during training.
Both policies condition actions on RGB observations, language instructions, and proprioception.
We train a separate pair of policies for each benchmark, using the same demonstration collection within each pair.
We report success rate (SR) using each benchmark's task-completion criterion.

On RoboTwin, both policies use 27,500 demonstrations across 50 tasks, with 50 clean and 500 randomized demonstrations per task.
A trajectory-level split assigns 27,225 demonstrations to training and 275 to validation; both policies are evaluated at step 29,365.
We also include the released RoboTwin-adapted $\pi_{0.5}$~\cite{pi052025,motuspi05robotwin} and MOTUS~\cite{bi2025motus,motusrobotwin} policies as additional baselines.
All four models are evaluated on all 50 tasks, with 100 valid episodes per task in each of six conditions: clean, randomized, and the four corruptions defined in \cref{sec:robotwin_experiments}.
For LIBERO, FastWAM and HOG-WAM use the same 2,000 demonstrations spanning 40 tasks from the Spatial, Object, Goal, and LIBERO-10 suites~\cite{liu2023libero}.
We evaluate both policies on the same 10,030 LIBERO Plus variants, using one rollout per variant from the first official initial state~\cite{fei2025liberoplus}.
This benchmark includes changes in background, lighting, sensor noise, camera viewpoint, language, object layout, and robot initialization.

\paragraph{Matched policy comparisons.}
Within each benchmark, FastWAM and HOG-WAM use the same action representation and prediction and execution horizons.
Both predict 32 actions; RoboTwin executes the first 24 joint-position targets, while LIBERO executes the first eight end-effector actions.
Both comparisons also match normalization statistics, effective batch size, optimization steps, and the ten-step action integration schedule.
The released RoboTwin baselines retain their respective execution horizons: 32 actions for $\pi_{0.5}$ and 16 for MOTUS.

\paragraph{Aggregation and sampling.}
RoboTwin conditions assign equal episode counts to all tasks, making the reported average equal to pooled success over the valid episodes.
The benchmark's expert-solvability criterion selects initial scenes; the same task populations and budgets are used across policies.
Scene seeds are unpaired across policies.
LIBERO Plus uses matched initial states, instructions, random seeds, and action budgets for each variant.
Its overall success rate pools variants, while category and suite tables group the same outcomes by perturbation and source task family.

\subsection{Visual-corruption testing on RoboTwin}
\label{sec:robotwin_experiments}
\paragraph{Controlled visual perturbations.}
RoboTwin 2.0~\cite{chen2025robotwin} supplies native clean and randomized scene settings.
We extend its clean-scene evaluation with four image corruptions: mixed noise, warm filtering, underexposure, and reduced contrast.
We use \textbf{RoboTwin-Plus} as shorthand for this visual-corruption test, which retains the original tasks, scenes, and success criteria.
Each operator acts on the policy's RGB inputs before visual encoding, leaving the task, language, and proprioceptive inputs unchanged.
The RoboTwin-Plus score averages success over the four corruption conditions.
Native clean and randomized results are reported separately.

Mixed noise combines additive Gaussian noise with standard deviation 10 in 8-bit intensity units and multiplicative speckle noise with standard deviation 0.05.
The operator rounds and clips the resulting values to the valid intensity range.
Warm filtering scales RGB channels by $(1.05,0.955,0.92)$, underexposure multiplies intensity by $2^{-0.5}$, and reduced contrast scales deviations from mid-gray by $0.80$.
These controlled changes isolate sensitivity to visual appearance within the existing task distribution.

\begin{table}[t]
    \centering
    \small
    \setlength{\tabcolsep}{2pt}
    \caption{RoboTwin success rates (\%). Each condition comprises 50 tasks with 100 episodes per task and model. RoboTwin-Plus averages the four corruption conditions. Bold marks the highest rate.}
    \label{tab:robotwin_summary}
    \begin{checkedtable}{\columnwidth}
    \begin{tabular}{@{}lrrrr@{}}
        \toprule
        Condition & $\pi_{0.5}$ & MOTUS & FastWAM & HOG-WAM \\
        \midrule
        Clean & 45.98 & 87.66 & \textbf{92.08} & 91.88 \\
        Randomized & 49.38 & 85.76 & 91.30 & \textbf{91.36} \\
        \midrule
        Mixed noise & 36.60 & 77.96 & 78.86 & \textbf{82.86} \\
        Warm filter & 44.62 & 86.28 & 89.16 & \textbf{89.96} \\
        Underexposure & 48.04 & 87.26 & 90.90 & \textbf{91.48} \\
        Reduced contrast & 47.00 & 87.52 & 89.14 & \textbf{90.00} \\
        \midrule
        RoboTwin-Plus mean & 44.07 & 84.76 & 87.02 & \textbf{88.58} \\
        \bottomrule
    \end{tabular}
    \end{checkedtable}
\end{table}

\paragraph{Visual robustness.}
HOG-WAM achieves the highest success among the four policies in every RoboTwin-Plus condition (\cref{tab:robotwin_summary}).
The gain concentrates under mixed noise, which causes the largest reduction in FastWAM success.
HOG-WAM also improves success under all three photometric corruptions.
Performance on native clean and randomized scenes remains comparable to FastWAM.
This pattern is consistent with learning to aggregate boundary evidence over neighborhoods through cellwise pooling and block normalization, reducing sensitivity to independent pixel noise.

\paragraph{Variation across tasks.}
Under mixed noise, HOG-WAM is higher than FastWAM on 28 of the 50 tasks, tied on 6, and lower on 16.
Gains are pronounced for block handover and placing bottles in a dustbin, while success is lower when placing a can in a basket or placing a shoe.
Under the photometric conditions, improved and lowered tasks are closer to balanced, in line with the small average differences.

\subsection{Robustness on LIBERO Plus}
\label{sec:libero_experiments}
\paragraph{Results by category and suite.}
The effect of HOG supervision on LIBERO Plus is category-specific (\cref{tab:libero_categories}).
HOG-WAM gains most under sensor noise and camera viewpoint changes and is lower under background and lighting changes; robot initialization, language, and object layout differ by less than 0.3 points.
A paired bootstrap over source tasks gives a 95\% interval for the difference that excludes zero only for sensor noise.
By source suite, HOG-WAM has higher success on Object and Spatial, with the largest gain in Object; FastWAM is slightly higher on Goal and LIBERO-10 (\cref{tab:libero_suites_main}).

\begin{table}[t]
\centering
\small
\setlength{\tabcolsep}{3pt}
\caption{LIBERO Plus success rates (\%) by perturbation category. FastWAM and HOG-WAM use the same 10,030 variants. $\Delta$ is HOG-WAM minus FastWAM in percentage points (pp), computed before rounding. Bold marks the higher rate.}
\label{tab:libero_categories}
\begin{checkedtable}{\columnwidth}
\begin{tabular}{@{}lrrr@{}}
\toprule
Category & FastWAM & HOG-WAM & $\Delta$ (pp) \\
\midrule
Background & \textbf{67.94} & 65.24 & $-2.70$ \\
Camera viewpoint & 64.60 & \textbf{68.67} & $+4.07$ \\
Robot initialization & \textbf{76.32} & 76.06 & $-0.26$ \\
Language & \textbf{73.78} & 73.65 & $-0.13$ \\
Lighting & \textbf{90.89} & 88.35 & $-2.54$ \\
Object layout & 76.39 & \textbf{76.46} & $+0.07$ \\
Sensor noise & 74.14 & \textbf{78.89} & $+4.75$ \\
\midrule
Overall & 74.49 & \textbf{75.26} & $+0.78$ \\
\bottomrule
\end{tabular}
\end{checkedtable}
\end{table}

\begin{table}[t]
\centering\small
\setlength{\tabcolsep}{4pt}
\caption{LIBERO Plus success rates (\%) by source suite. FastWAM and HOG-WAM use the same variants. Bold marks the higher rate within each suite.}
\label{tab:libero_suites_main}
\begin{checkedtable}{\columnwidth}
\begin{tabular}{@{}lrr@{}}
\toprule
Suite & FastWAM & HOG-WAM \\
\midrule
Spatial & 83.47 & \textbf{84.47} \\
Object & 80.78 & \textbf{84.00} \\
Goal & \textbf{71.40} & 70.51 \\
LIBERO-10 & \textbf{62.80} & 62.64 \\
\bottomrule
\end{tabular}\end{checkedtable}
\end{table}

\begin{figure*}[t]
    \centering
    \includegraphics[width=.78\textwidth]{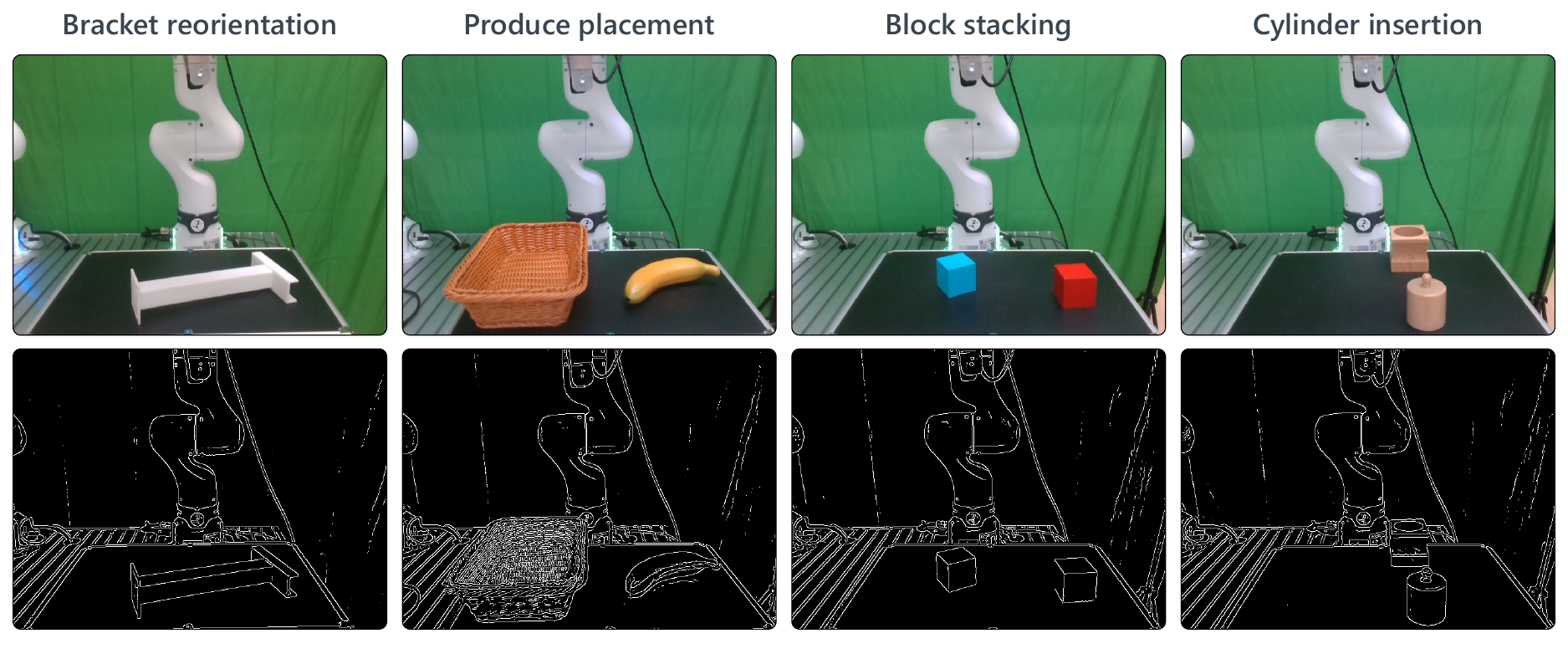}
    \caption{Franka manipulation tasks. Columns show bracket reorientation, produce placement, block stacking, and cylinder insertion. The upper row contains initial main-camera RGB observations from recorded demonstrations; the lower row shows gradients of the same images. These monochrome displays illustrate local boundaries; training uses the HOG-derived raster defined in \cref{sec:hog_targets}.}
    \label{fig:franka_tasks}
\end{figure*}

\subsection{Real-world Franka manipulation}
\label{sec:real_robot}
We evaluate FastWAM and HOG-WAM on four Franka tasks: bracket reorientation, produce placement, block stacking, and cylinder insertion (\cref{fig:franka_tasks}).
The collection contains 1,000 demonstrations, with 200 for each task except produce placement, which has 400.
Both policies use the same demonstrations, camera views, language instructions, and joint-position action representation.
We evaluate both models at training step 30,000.

The tasks require distinct final spatial relations: an inverted bracket, produce contained in a basket, a stable block stack, and a seated cylinder.
The policies observe a main-camera view and a wrist-camera view, together with measured joint positions and gripper width.
Demonstrations are recorded at 30 Hz and divided by trajectory into 900 training and 100 validation examples.
Both models use normalization statistics from the training trajectories and predict absolute joint targets with a binary gripper command.
Each trial lasts at most 200 steps and terminates early on success or a failure/error.
After every trial, the robot is reset using the reset command.

Each task includes 25 trials per model under clean observations and 25 under the RoboTwin-Plus mixed-noise condition, giving 400 physical trials overall.
\Cref{tab:franka_results} reports task success rates and their macro-average.
\begin{table}[!ht]
\centering\small
\setlength{\tabcolsep}{3pt}
\caption{Franka success rates (\%). Each task has 25 trials per model and condition. The noisy condition applies the RoboTwin-Plus mixed-noise operator. F: FastWAM; H: HOG-WAM. Bold marks the higher rate within each condition; ties remain unbolded.}
\label{tab:franka_results}
\begin{checkedtable}{\columnwidth}
\begin{tabular}{@{}lrrrr@{}}
\toprule
& \multicolumn{2}{c}{Clean} & \multicolumn{2}{c}{Noisy} \\
\cmidrule(lr){2-3}\cmidrule(l){4-5}
Task & F & H & F & H \\
\midrule
Bracket reorientation & 40 & \textbf{48} & 12 & \textbf{20} \\
Produce placement & 60 & \textbf{68} & 32 & \textbf{40} \\
Block stacking & 32 & \textbf{80} & 24 & \textbf{36} \\
Cylinder insertion & 28 & \textbf{40} & 0 & 0 \\
\midrule
Macro-average & 40 & \textbf{59} & 17 & \textbf{24} \\
\bottomrule
\end{tabular}\end{checkedtable}
\end{table}

HOG-WAM achieves higher success on all four tasks under clean observations.
The gain concentrates in block stacking; the other three tasks improve by two or three successes out of 25 trials.
Under noise, it retains an advantage in reorientation, placement, and stacking, while neither policy completes cylinder insertion.

\subsection{Ablation study}
\label{sec:ablation_experiments}
We compare FastWAM, RGB-to-HOG, and HOG-WAM to distinguish auxiliary structural supervision from replacing RGB inputs and targets.
RGB-to-HOG converts observations to HOG during training and inference; its visual DiT predicts future HOG, while its action DiT predicts actions.
HOG-WAM retains RGB observations and RGB prediction, adding a separate HOG DiT for auxiliary supervision.
All configurations share the 2,000 LIBERO demonstrations, effective batch size, and optimization steps.
Evaluation uses the same 10,030 LIBERO Plus variants, initial states, and rollout settings.

\par\medskip
\noindent\begin{minipage}{\columnwidth}
    \centering
    \small
    \setlength{\tabcolsep}{3pt}
    \captionof{table}{Ablation of complementary supervision and RGB replacement, evaluated on LIBERO Plus. Input specifies the observation representation; targets specify future visual supervision. Each model is evaluated on the same 10,030 variants. SR denotes success rate; bold marks the highest rate.}
    \label{tab:libero_ablation}
    \begin{checkedtable}{\columnwidth}
    \begin{tabular}{@{}lccr@{}}
        \toprule
        Model & Input & Targets & SR (\%) \\
        \midrule
        FastWAM & RGB & RGB & 74.49 \\
        RGB-to-HOG & HOG & HOG & 66.80 \\
        HOG-WAM & RGB & RGB + HOG & \textbf{75.26} \\
        \bottomrule
    \end{tabular}
    \end{checkedtable}
\end{minipage}
\par\medskip

Auxiliary supervision yields the highest overall success among the three configurations (\cref{tab:libero_ablation}).
Both RGB-input policies achieve higher success than RGB-to-HOG under viewpoint, noise, language, initial-state, and object-layout changes (\cref{tab:libero_ablation_categories_main}).
RGB-to-HOG leads under background and lighting changes, the two categories whose perturbations are confined to appearance that its input discards.

\par\medskip
\noindent\begin{minipage}{\columnwidth}
\centering\small
\setlength{\tabcolsep}{3pt}
\captionof{table}{Representation ablation by perturbation category on LIBERO Plus. Entries are success rates (\%) on the same variants for all three models. Bold marks the highest rate within each category.}
\label{tab:libero_ablation_categories_main}
\begin{checkedtable}{\columnwidth}
\begin{tabular}{@{}lrrr@{}}
\toprule
Category & FastWAM & RGB-to-HOG & HOG-WAM \\
\midrule
Background & 67.94 & \textbf{75.65} & 65.24 \\
Camera viewpoint & 64.60 & 43.21 & \textbf{68.67} \\
Robot initialization & \textbf{76.32} & 71.35 & 76.06 \\
Language & \textbf{73.78} & 60.57 & 73.65 \\
Lighting & 90.89 & \textbf{91.77} & 88.35 \\
Object layout & 76.39 & 71.15 & \textbf{76.46} \\
Sensor noise & 74.14 & 64.02 & \textbf{78.89} \\
\bottomrule
\end{tabular}
\end{checkedtable}
\end{minipage}
\par\medskip

The replacement model changes both the observation and the visual prediction problem.
Its compressed orientation statistics suppress appearance variation but also remove chromatic distinctions that can identify an instructed object.
HOG-WAM retains these cues in RGB, and its response to background and lighting changes stays close to the RGB-input baseline.
Auxiliary HOG supervision improves success under sensor noise and viewpoint changes while preserving the cues that RGB-to-HOG loses in the remaining categories.
Under sensor noise the ordering reverses: HOG computed from corrupted observations is the weakest input, whereas HOG computed from clean frames is an effective target.
The comparison favors separating the information presented to the policy from the target used to shape its representation.

Both RGB-input policies outperform RGB-to-HOG in every source suite (\cref{tab:libero_ablation_suites}), indicating that RGB input is needed across task families.

\par\medskip
\noindent\begin{minipage}{\columnwidth}
\centering\small
\setlength{\tabcolsep}{3pt}
\captionof{table}{Representation ablation by source suite on LIBERO Plus. Entries report success rates (\%); all configurations use the same variants. Bold marks the highest rate within each suite.}
\label{tab:libero_ablation_suites}
\begin{checkedtable}{\columnwidth}
\begin{tabular}{@{}lrrr@{}}
\toprule
Suite & FastWAM & RGB-to-HOG & HOG-WAM \\
\midrule
Spatial & 83.47 & 78.02 & \textbf{84.47} \\
Object & 80.78 & 72.12 & \textbf{84.00} \\
Goal & \textbf{71.40} & 64.84 & 70.51 \\
LIBERO-10 & \textbf{62.80} & 52.80 & 62.64 \\
\bottomrule
\end{tabular}
\end{checkedtable}
\end{minipage}
\par\medskip

\section{Discussion}
\label{sec:discussion}
The experiments favor assigning gradients and RGB different roles: gradients structure the learning objective, while RGB preserves the information available to the policy.
The design combines MaskFeat's use of HOG as a predictive target~\cite{wei2022maskfeat} with a removable structural readout in the manner of WAM4D~\cite{li2026wam4d}, obtaining the supervision from deterministic local image statistics.
Its gains concentrate under sensor noise and viewpoint changes.

The physical results extend the evidence beyond rendered observations, with larger absolute gains than the simulation averages.
Each domain uses its own demonstrations, so these results establish effectiveness in both simulation and physical manipulation.
The HOG DiT is removed at inference, so both policies execute the same architecture with the same parameter count, and the benefit resides in the trained RGB and action weights.

\section{Conclusion}
\label{sec:conclusion}
HOG-WAM adds deterministic future-gradient supervision to predictive policy learning through an auxiliary DiT module and asymmetric attention.
The resulting policy improves average success on RoboTwin-Plus, LIBERO Plus, and four Franka tasks while preserving the FastWAM inference architecture.
The representation ablation supports retaining RGB observations and using structural prediction as complementary supervision; the gain is clearest under sensor noise.

\FloatBarrier
\section*{LLM Usage Statement}
OpenAI Codex assisted with drafting and revising the abstract, introduction, related work, method, and experimental descriptions; preparing and inspecting implementation and configuration code; coordinating and diagnosing remote training and evaluation workflows; developing code for result aggregation and consistency checks; preparing the LaTeX layout; and assisting with diagram layouts, editable vector and presentation figures, and image-processing scripts for gradient visualizations.
OpenAI Codex also assisted with scripts for composing the recorded demonstration videos and preparing the submission files.
Devin assisted with language polishing of the manuscript.
The photographic panels are recorded observations; gradient displays are computed from those images.
The reported success counts are calculated from recorded evaluation outcomes.
The authors are responsible for reviewing and verifying all AI-assisted material and retain full responsibility for the scientific claims, source attribution, experimental validity, and submitted content.

\section*{Acknowledgments}
This research was supported in part with Cloud TPUs from Google's TPU Research Cloud (TRC).

{\small
\bibliographystyle{ieeenat_fullname}
\bibliography{references}
}
\clearpage
\onecolumn
\raggedbottom
\appendix
\setcounter{table}{0}
\setcounter{figure}{0}
\setcounter{equation}{0}
\renewcommand{\theequation}{S\arabic{equation}}
\renewcommand{\theHtable}{supp.\arabic{table}}
\renewcommand{\theHfigure}{supp.\arabic{figure}}
\renewcommand{\theHequation}{supp.\arabic{equation}}
\begin{center}\Large\textbf{Supplementary Material}\end{center}

\renewcommand{\thetable}{S\arabic{table}}
\renewcommand{\thefigure}{S\arabic{figure}}
\section{LIBERO Plus results by source suite}
\Cref{tab:libero_suites} groups the evaluations by source suite.
The suite scores summarize the same variants used in the category breakdown, and the overall score pools all variants.

\par\medskip
\noindent\begin{minipage}{\textwidth}

\centering
\small
\setlength{\tabcolsep}{3.5pt}
\captionof{table}{LIBERO Plus results grouped by the source suite of each perturbed task. Each model cell reports successes / success rate (SR, \%). $\Delta$ is HOG-WAM minus FastWAM in counts or percentage points (pp), computed before rounding. Bold marks the higher rate. The overall score pools all evaluated variants.}
\label{tab:libero_suites}
\begin{checkedtable}{\textwidth}
\begin{tabular}{@{}lrrrrr@{}}
\toprule
Suite & Trials & \multicolumn{2}{c}{Successes / SR (\%)} & \multicolumn{2}{c}{$\Delta$} \\
\cmidrule(lr){3-4}\cmidrule(l){5-6}
& $n$ & FastWAM & HOG-WAM & Count & pp \\
\midrule
Spatial & 2,402 & 2,005 / 83.47 & \textbf{2,029 / 84.47} & $+24$ & $+1.00$ \\
Object & 2,518 & 2,034 / 80.78 & \textbf{2,115 / 84.00} & $+81$ & $+3.22$ \\
Goal & 2,591 & \textbf{1,850 / 71.40} & 1,827 / 70.51 & $-23$ & $-0.89$ \\
LIBERO-10 & 2,519 & \textbf{1,582 / 62.80} & 1,578 / 62.64 & $-4$ & $-0.16$ \\
\midrule
Overall & 10,030 & 7,471 / 74.49 & \textbf{7,549 / 75.26} & $+78$ & $+0.78$ \\
\bottomrule
\end{tabular}
\end{checkedtable}

\end{minipage}
\par\medskip

\section{Training and evaluation details}
\label{supp:sec:evaluation}
\subsection{Training configurations}
The RoboTwin collection contains 27,500 demonstrations from 50 tasks: 50 clean and 500 randomized demonstrations per task.
Both policies use the same trajectory-level split of 27,225 training and 275 validation demonstrations and eight independently noised action copies per visual sample.
Training uses AdamW with peak learning rate $10^{-4}$, weight decay $10^{-2}$, and effective batch size 1,024 for five epochs; evaluation uses the final step-29,365 checkpoints.
The learning rate warms up for 1,468 steps and then decays by a cosine schedule to $10^{-6}$.
The optimizer uses $(\beta_1,\beta_2)=(0.9,0.95)$, $\epsilon=10^{-8}$, and gradient-norm clipping at 1.
\Cref{supp:tab:training} summarizes the benchmark-specific settings.
For LIBERO, FastWAM and HOG-WAM use the same 40-task, 2,000-demonstration collection, normalization statistics, effective batch size of 256, and action-copy count $\tau=8$.
Training uses a peak learning rate of $10^{-4}$, weight decay of $10^{-2}$, and ten epochs.
We evaluate the final checkpoints at step 13,200.
The models with visual and action DiT modules assign unit weights to their visual and action objectives; HOG-WAM additionally assigns unit weight to its HOG objective.
The auxiliary HOG DiT module increases training computation and parameter count.
\par\medskip
\noindent\begin{minipage}{\textwidth}
\centering\small
\captionof{table}{Training settings shared by FastWAM and HOG-WAM within each benchmark. Dimensions are height $\times$ width. The Franka learning-rate schedule is specified in \cref{supp:real_robot}.}
\label{supp:tab:training}
\begin{checkedtable}{\textwidth}
\begin{tabular}{@{}lccc@{}}
\toprule
Setting & RoboTwin & LIBERO & Franka \\
\midrule
Task count & 50 & 40 & 4 \\
Demonstration collection & 27,500 & 2,000 & 1,000 \\
Training / validation trajectories & 27,225 / 275 & 2,000 / 0 & 900 / 100 \\
RGB canvas & $384\times320$ & $224\times448$ & $384\times320$ \\
Camera views & Main + two wrists & Agent + wrist & Main + wrist \\
Action dimension $d_a$ & 14 & 7 & 8 \\
Predicted actions $K$ & 32 & 32 & 32 \\
Effective batch size & 1,024 & 256 & 128 \\
Action copies $\tau$ & 8 & 8 & 8 \\
Peak learning rate & $10^{-4}$ & $10^{-4}$ & $2.5\times10^{-5}$ \\
Evaluated training step & 29,365 & 13,200 & 30,000 \\
\bottomrule
\end{tabular}
\end{checkedtable}
\end{minipage}
\par\medskip

Each RoboTwin visual clip contains the current observation and eight future observations sampled every four action steps, spanning 32 actions.
The main camera occupies the upper $256\times320$ pixels of the $384\times320$ canvas; two $128\times160$ wrist views form the lower row.
LIBERO concatenates its agent and wrist views into a $224\times448$ canvas.
All canvas dimensions are height $\times$ width.

\subsection{RoboTwin visual-corruption extension}
\Cref{fig:robotwin_protocol} illustrates the four observation corruptions.

\par\medskip
\noindent\begin{minipage}{\textwidth}

    \centering
    \includegraphics[width=.73\textwidth]{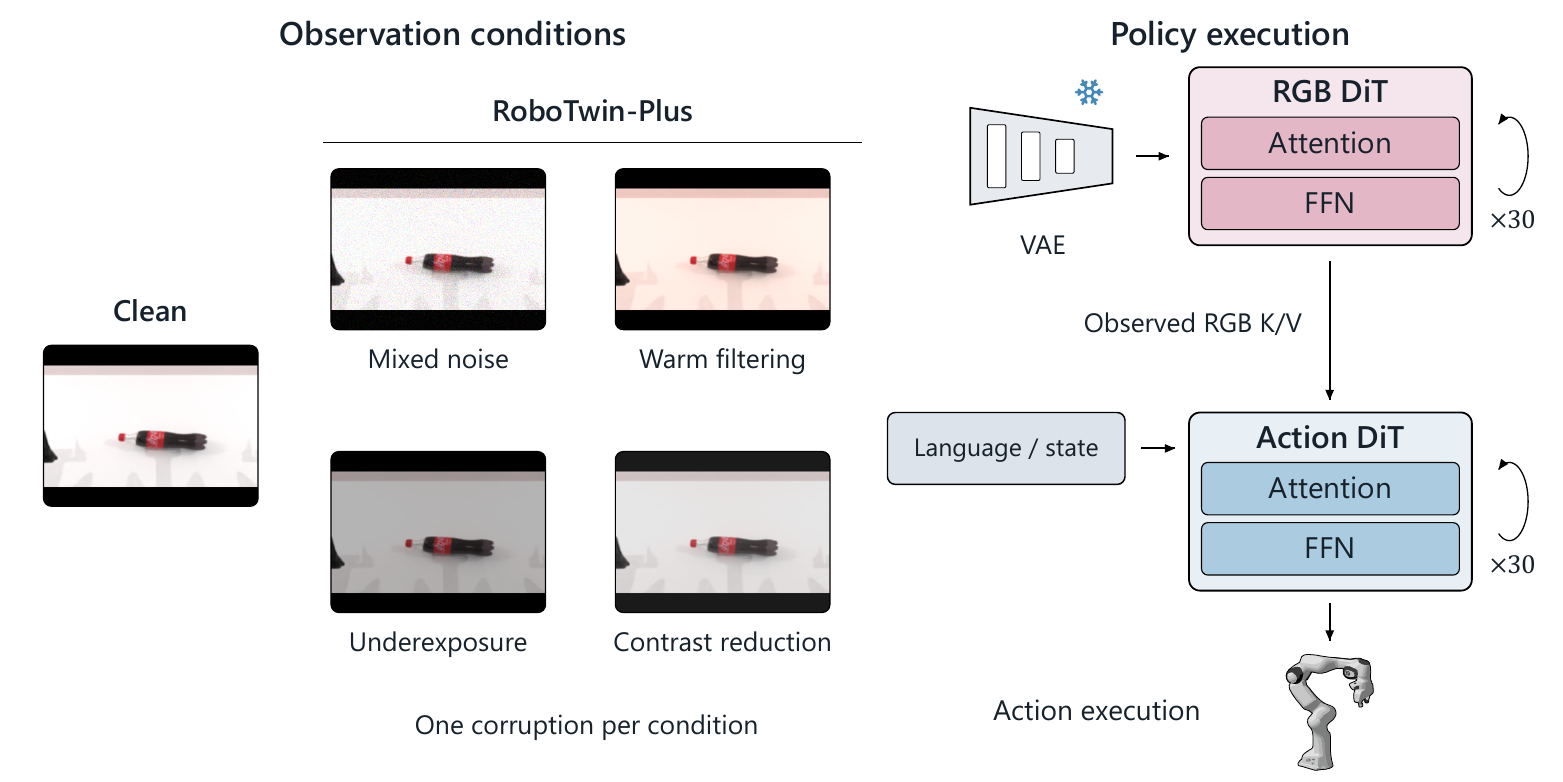}
    \captionof{figure}{RoboTwin visual-corruption extension (RoboTwin-Plus). Clean observations are shown separately from the four corruption conditions. A single corruption operator modifies the RGB observation before visual encoding, preserving the manipulation objective and nonvisual conditioning. Mixed noise, warm filtering, underexposure, and reduced contrast probe complementary forms of visual distribution shift. Each completed evaluation comprises 50 tasks with 100 valid episodes per task, model, and condition; the RoboTwin-Plus score averages success across the four conditions.}
    \label{fig:robotwin_protocol}

\end{minipage}
\par\medskip

For FastWAM and HOG-WAM, perturbations act on the composed 8-bit RGB canvas before visual encoding.
The additional baselines retain their native camera layouts; $\pi_{0.5}$ receives three separately resized and padded $224\times224$ RGB views.
The same intensity operators and numerical parameters are applied to each policy's image inputs.
The mixed-noise condition adds Gaussian noise with standard deviation 10 intensity levels, then applies multiplicative speckle noise with standard deviation 0.05:
\begin{equation}
    \begin{aligned}
    \widetilde I &= Q\!\left((I+\epsilon_g)\odot(1+\epsilon_s)\right),\\
    \epsilon_g&\sim\mathcal N(0,10^2),\quad
    \epsilon_s\sim\mathcal N(0,0.05^2).
    \end{aligned}
\end{equation}
Here, $Q$ rounds and clips values to $[0,255]$, and $\odot$ denotes elementwise multiplication.
Gaussian and speckle samples are independent across pixels and channels and are redrawn at each observation update.
A deterministic seed combines the experiment seed, episode index, and observation-update index.
The photometric operators apply channel-wise warm-color scaling, an exposure multiplier, or intensity contraction toward mid-gray:
\begin{equation}
\begin{aligned}
    \mathcal C_{\mathrm{warm}}(I)
        &=Q\!\left(I\odot[(1-\alpha)\mathbf1+\alpha\mathbf t]\right),\\
    \mathcal C_{\mathrm{under}}(I)
        &=Q\!\left(2^e I\right),\\
    \mathcal C_{\mathrm{contrast}}(I)
        &=Q\!\left(k(I-127.5)+127.5\right).
\end{aligned}
\end{equation}
The warm tint is $\mathbf t=(1.20,0.82,0.68)$ with strength $\alpha=0.25$, giving channel multipliers $(1.05,0.955,0.92)$.
Underexposure uses $e=-0.5$, and reduced contrast uses $k=0.80$.
These parameters are fixed throughout each condition; photometric operators are applied separately from mixed noise.

\subsection{Evaluation sampling and action prediction}
For RoboTwin, each completed model--condition evaluation contains 100 valid episodes for each of 50 tasks.
Episodes interrupted by simulation errors are excluded from the success-rate denominator.
Initial scenes are selected by the benchmark's expert-solvability criterion.
The comparisons use the same task populations and episode budgets.
The expert-solvability criterion can yield different scene seeds across policies, so initial scenes are unpaired.

FastWAM and HOG-WAM predict 32 absolute joint-position actions and execute the first 24 before replanning.
Their four RoboTwin-Plus conditions each contain 5,000 episodes per model.
The equally weighted mean across conditions is also the pooled success rate over these episodes.
Both policies are evaluated in all six conditions with these episode budgets.
The additional $\pi_{0.5}$ and MOTUS baselines are described below.

For LIBERO Plus, FastWAM and HOG-WAM use the same 10,030 variants, instructions, initial states, random seeds, and action-step budgets.
Each policy predicts 32 actions with ten flow-integration steps, executes the first eight, and replans synchronously.
The overall score pools all variants; category and suite breakdowns group these same outcomes in two ways.
Each variant uses its first official initial state.

\subsection{Additional \texorpdfstring{$\pi_{0.5}$}{pi0.5} baseline}
We evaluate the publicly released \texttt{motus-robotics/pi0.5\_robotwin2} model~\cite{pi052025,motuspi05robotwin}, using the weights at training step 40,000 and normalization statistics for the combined clean and randomized training data.
The policy receives measured joint positions, predicts 32 actions with ten denoising steps, and executes all 32 before obtaining the next observation.
This measured-joint-state protocol defines our reproduction setting, with a longer execution horizon than the 24 actions used by FastWAM and HOG-WAM.
The release metadata describes constructing states from actions; the source for that preprocessing is unavailable.

Each of the six conditions covers all 50 tasks with 100 episodes per task.
The supplementary task tables report the complete per-task and aggregate success counts.

\subsection{Additional MOTUS baseline}
We evaluate the publicly released \texttt{motus-robotics/Motus\_robotwin2} model~\cite{bi2025motus,motusrobotwin} at training step 40,000.
The policy receives three RGB camera views, language instructions, and measured joint positions, and uses the released action-normalization statistics.
It predicts and executes 16 actions per observation with ten denoising steps.
The evaluation covers the same six conditions, with 100 valid episodes for each of the 50 tasks per condition.

\subsection{Evaluation coverage}
The reported results evaluate one trained checkpoint per model and benchmark at the specified corruption severities.
The RoboTwin photometric comparisons use the parameters specified above, and mixed noise follows the prescribed Gaussian-plus-speckle distribution.
LIBERO Plus contributes one rollout per evaluated variant.
These choices define the sampling protocol for all reported aggregate and category-level results.
Where differences are tabulated for LIBERO Plus, pp denotes percentage points and values are computed before rounding.

\subsection{Uncertainty across source tasks}
The simulation comparisons use fixed trained checkpoints.
To assess sensitivity to task composition, we construct percentile intervals from 20,000 bootstrap resamples of source tasks, using random seed 20261117.
All outcomes belonging to a sampled task remain together, preserving dependencies among its perturbation variants and the policy comparison.

For LIBERO Plus, we verify the pairing of all 10,030 variants by suite, task identifier, instruction, initial-state seed, and action budget.
Within each of the four source suites, we sample ten base tasks with replacement and retain all variants of each selected task for both models.
We recalculate the pooled success-rate difference in each resample, so tasks contribute in proportion to their variant counts as in the reported benchmark score.
\Cref{supp:tab:libero_intervals} reports the resulting pointwise 95\% intervals.
The overall interval spans zero, whereas the sensor-noise interval is positive; the category analysis identifies where the gain is most consistent across source tasks.
These intervals describe task-sampling uncertainty conditional on the evaluated policies; repeated training and repeated rollouts of a fixed variant are separate sources of variation.

\par\medskip
\noindent\begin{minipage}{\textwidth}
\centering\small
\captionof{table}{Paired task-cluster bootstrap analysis on LIBERO Plus. $\Delta$ is HOG-WAM minus FastWAM in percentage points (pp). Intervals preserve paired variant outcomes within source tasks and are unadjusted for multiple comparisons.}
\label{supp:tab:libero_intervals}
\begin{checkedtable}{\textwidth}
\begin{tabular}{@{}lrr@{}}
\toprule
Category & $\Delta$ (pp) & 95\% interval (pp) \\
\midrule
Overall & $+0.78$ & $[-1.14, +2.67]$ \\
Background & $-2.70$ & $[-6.75, +1.28]$ \\
Camera viewpoint & $+4.07$ & $[-0.14, +8.16]$ \\
Robot initialization & $-0.26$ & $[-3.28, +2.66]$ \\
Language & $-0.13$ & $[-4.98, +4.45]$ \\
Lighting & $-2.54$ & $[-7.65, +1.30]$ \\
Object layout & $+0.07$ & $[-2.11, +2.08]$ \\
Sensor noise & $+4.75$ & $[+0.88, +8.74]$ \\
\bottomrule
\end{tabular}
\end{checkedtable}
\end{minipage}
\par\medskip

For RoboTwin, we sample 50 task clusters with replacement and recompute each policy's absolute mean success rate (\cref{supp:tab:robotwin_intervals}).
Each cluster retains the 100 episodes for its task and policy; initial scenes remain unpaired across policies.
The resampling therefore measures sensitivity to the task mix without imposing episode-level pairing.

\par\medskip
\noindent\begin{minipage}{\textwidth}
\centering\small
\captionof{table}{Pointwise 95\% task-cluster bootstrap intervals for RoboTwin success rates (\%). Each condition uses 50 source tasks with 100 episodes per task and policy. Intervals summarize variation in the resampled task mix at fixed checkpoints.}
\label{supp:tab:robotwin_intervals}
\begin{checkedtable}{\textwidth}
\begin{tabular}{@{}lcccc@{}}
\toprule
Condition & $\pi_{0.5}$ & MOTUS & FastWAM & HOG-WAM \\
\midrule
Clean & [39.44, 52.76] & [83.28, 91.40] & [88.88, 94.80] & [88.50, 94.80] \\
Randomized & [43.02, 55.90] & [81.72, 89.32] & [88.46, 93.80] & [88.34, 94.02] \\
Mixed noise & [29.76, 43.90] & [71.88, 83.38] & [72.06, 85.00] & [77.88, 87.48] \\
Warm filter & [37.98, 51.50] & [81.46, 90.44] & [85.40, 92.44] & [85.88, 93.40] \\
Underexposure & [41.56, 54.64] & [82.90, 91.02] & [87.32, 93.94] & [88.46, 94.18] \\
Reduced contrast & [40.40, 53.78] & [82.86, 91.44] & [85.26, 92.56] & [85.76, 93.54] \\
\bottomrule
\end{tabular}
\end{checkedtable}
\end{minipage}
\par\medskip

\section{Additional method details}
\subsection{Deterministic HOG targets}
\label{supp:sec:hog_targets}

We apply a deterministic transform $\mathcal{H}$ independently to each frame of the composed RGB clip.
After mapping RGB values to $[0,1]$, we compute the grayscale intensity
\begin{equation}
    Y=0.2989R+0.5870G+0.1140B.
\end{equation}
Centered finite differences with replicated boundaries yield horizontal and vertical gradients $g_x$ and $g_y$.
Their magnitude and unsigned orientation are
\begin{equation}
    m=\sqrt{g_x^2+g_y^2},\qquad
    \phi=\operatorname{atan2}(g_y,g_x)\bmod\pi.
\end{equation}
Within each $8\times8$ cell, gradient magnitudes are accumulated into nine orientation bins using linear interpolation between neighboring bins.
We apply L2-Hys normalization over overlapping $2\times2$-cell blocks by normalizing, clipping at $0.2$, and renormalizing with a small positive stabilizer.
Contributions from overlapping blocks are averaged for each cell.

To match the video VAE input, we rasterize the normalized cell histograms $h_{qk}$ into three channels.
Here, $q$ indexes a cell and $k\in\{0,\ldots,8\}$ indexes an orientation bin.
With $\beta_k=(k+\tfrac12)\pi/9$, define
\begin{equation}
\begin{aligned}
    e_q&=\sum_k h_{qk}, &
    c_q&=\sum_k h_{qk}\cos(2\beta_k),\\
    d_q&=\sum_k h_{qk}\sin(2\beta_k), &
    b&=\max(\max_q e_q,\varepsilon).
\end{aligned}
\end{equation}
The cell raster $r_q=(2e_q/b-1,c_q/b,d_q/b)$ is clipped to $[-1,1]$ and repeated across the pixels in the cell.
The doubled angle provides a consistent representation of unsigned orientation.
The result is a \emph{HOG-derived three-channel image} that summarizes the nine-bin cell histograms.
It is encoded using the same frozen VAE weights and latent convention as the RGB stream:
\begin{equation}
    Z^h=\mathcal{E}\bigl(\mathcal{H}(I_{0:T})\bigr).
\end{equation}
Targets are derived from recorded demonstration frames, including ground-truth future frames.
The transform has no trainable parameters and requires neither object annotations nor a pretrained model for target generation.

\subsection{Coupled diffusion transformer architecture}
\label{supp:sec:architecture}

During training, HOG-WAM contains an RGB DiT module, a HOG DiT module, and an action DiT module.
The RGB and HOG DiTs share their architecture and token geometry.
Both are initialized from the pretrained RGB DiT and retain separate trainable parameters.
The action DiT and its initialization follow the matched FastWAM configuration, with benchmark-specific input and output projections.
\Cref{supp:tab:architecture} summarizes the backbone configuration.

\par\medskip
\noindent\begin{minipage}{\textwidth}

    \centering
    \small
    \captionof{table}{HOG-WAM backbone configuration. Visual inputs are VAE latents divided into $1\times2\times2$ spatiotemporal patches; action inputs are normalized coordinates. A common attention-head geometry connects visual and action DiTs with different hidden widths through mixed attention. The frozen VAE and text encoder are separate from these DiTs. Here, $d_a$ is the action dimension: 14 for RoboTwin, 7 for LIBERO, and 8 for Franka.}
    \label{supp:tab:architecture}
\begin{checkedtable}{\textwidth}
\setlength{\tabcolsep}{3pt}
\begin{tabular}{@{}lccc@{}}
        \toprule
        Component & RGB & HOG & Action \\
        \midrule
        Transformer layers & 30 & 30 & 30 \\
        Hidden width & 3072 & 3072 & 1024 \\
        Feed-forward width & 14336 & 14336 & 4096 \\
        Attention heads & 24 & 24 & 24 \\
        Head dimension & 128 & 128 & 128 \\
        Input channels/dim. & 48 & 48 & $d_a$ \\
        Used at inference & Yes & No & Yes \\
        \bottomrule
    \end{tabular}
\end{checkedtable}

\end{minipage}
\par\medskip

Each DiT embeds its noisy state and flow time using modality-specific transformer parameters.
At each layer, queries, keys, and values are projected to a common head geometry and combined through masked mixed attention.
The attention output is split into three streams, each with its own output projection, contextual cross-attention, and feed-forward network.
All streams receive the instruction and projected proprioceptive context.
The RGB and HOG output heads predict latent velocities, while the action head predicts velocities in normalized action coordinates.
Masked attention couples the three DiT modules during training.

\subsection{Asymmetric attention and removable auxiliary supervision}
\label{supp:sec:attention}

The attention diagram in the main paper depicts the visibility pattern at each coupled DiT layer.
RGB queries attend to RGB keys, and HOG queries attend to both RGB and HOG keys.
Both visual streams follow the \emph{first-frame-causal} rule: observed-frame queries attend only to observed-frame keys, whereas future queries attend to the observed frame and all noisy future frames.
Future visual tokens can therefore interact in both temporal directions.
Action queries attend to observed RGB keys and to keys from the same action copy.
The mask isolates independently sampled action copies and excludes HOG features from action conditioning.

This structure has two properties.
First, RGB and action forward computations are independent of HOG states, allowing inference to use the RGB and action DiT modules alone.
Observed RGB features depend on the current observation, keeping action conditioning consistent between training and inference.
Second, HOG predictions depend on RGB features through the permitted HOG-to-RGB attention connections, allowing the auxiliary loss to update the RGB backbone through backpropagation.
For fixed parameters and conditioning, these dependencies can be expressed as
\begin{equation}
    \frac{\partial\widehat{u}^{a}}{\partial Z^h}=0,
    \qquad
    \nabla_{\theta_v}\mathcal{L}_h\not\equiv0,
\end{equation}
where $\widehat{u}^{a}$ is the action velocity prediction and $\theta_v$ denotes RGB DiT parameters.
The first relation expresses forward independence; the second identifies the gradient path from HOG prediction to the RGB backbone.
In the standard configuration, the HOG DiT receives its clean first latent frame during training, but RGB and action queries cannot attend to this conditioning frame.

\subsection{Joint flow-matching training}
\label{supp:sec:training}

We retain the flow-matching formulation~\cite{lipman2023flowmatching} of the baseline.
For each modality $m\in\{v,h,a\}$, let $X^v=Z^v$, $X^h=Z^h$, and $X^a=A$ be the clean training states.
Gaussian noise $\epsilon^m$ and flow times are sampled independently for the three streams.
Using $\sigma_m\in[0,1]$ to denote the resulting noise level, the corrupted state and target velocity are
\begin{equation}
    \widetilde{X}^m=(1-\sigma_m)X^m+\sigma_m\epsilon^m,
    \qquad u^m=\epsilon^m-X^m.
\end{equation}
The baseline time sampler applies a shift to a uniform draw, $\sigma=\eta r/[1+(\eta-1)r]$, where $r\sim\mathcal{U}(0,1)$ and $\eta=5$.
After corruption, the first RGB and HOG latent frames are replaced by their clean conditioning values.
Their token-wise flow times are also set to zero, matching the clean observed-frame computation used at inference.
Predictions for these first frames are excluded from the corresponding visual losses; only future latent frames are supervised.

The model jointly predicts $\widehat{u}^v$, $\widehat{u}^h$, and $\widehat{u}^a$ using the mixed-attention backbone.
For each modality, the loss is a weighted velocity-prediction error,
\begin{equation}
    \mathcal{L}_m=\mathbb{E}\left[
        w(\sigma_m)
        \left\|\widehat{u}^m-u^m\right\|_{\mathrm{valid}}^2
    \right],
\end{equation}
where $\|\cdot\|_{\mathrm{valid}}^2$ averages over feature dimensions and valid target positions, excluding padded observations and actions.
We retain the baseline time-dependent weight,
\begin{equation}
    w(\sigma)=
    \frac{\exp[-2(\sigma-\tfrac12)^2]-\exp(-\tfrac12)}
         {0.1894955536+10^{-10}},
\end{equation}
and combine the objectives as
\begin{equation}
    \mathcal{L}=
    \lambda_v\mathcal{L}_v+
    \lambda_h\mathcal{L}_h+
    \lambda_a\mathcal{L}_a.
    \label{supp:eq:joint_loss}
\end{equation}
The configuration with three DiT modules sets all three coefficients to one.
The HOG term supervises velocity prediction in the frozen video latent space.

Our full-training configurations use eight copies of the same action chunk per visual sample, each with independently sampled noise and flow time.
The copies are packed together; each attends to the observed RGB frame and to itself, but not to other copies, and the action loss is averaged across them.
The setting $\tau=8$ denotes this Monte Carlo replication of the action objective.
The packed copies share a single RGB/HOG visual prediction, and the baseline uses the same number of action copies.

For full-parameter training, the RGB DiT, HOG DiT, action DiT, and proprioceptive projection are optimized, while the video VAE and text encoder remain frozen.
The HOG transform itself has no parameters to optimize.
The frozen encoders preserve the pretrained latent representations, while the DiT modules adapt to the joint prediction objectives.

\par\medskip
\noindent\begin{minipage}{\textwidth}
\subsection{Inference}
\label{supp:sec:deployment}
\Cref{fig:deployment} shows action generation from the current observation.

    \centering
    \includegraphics[width=.73\textwidth]{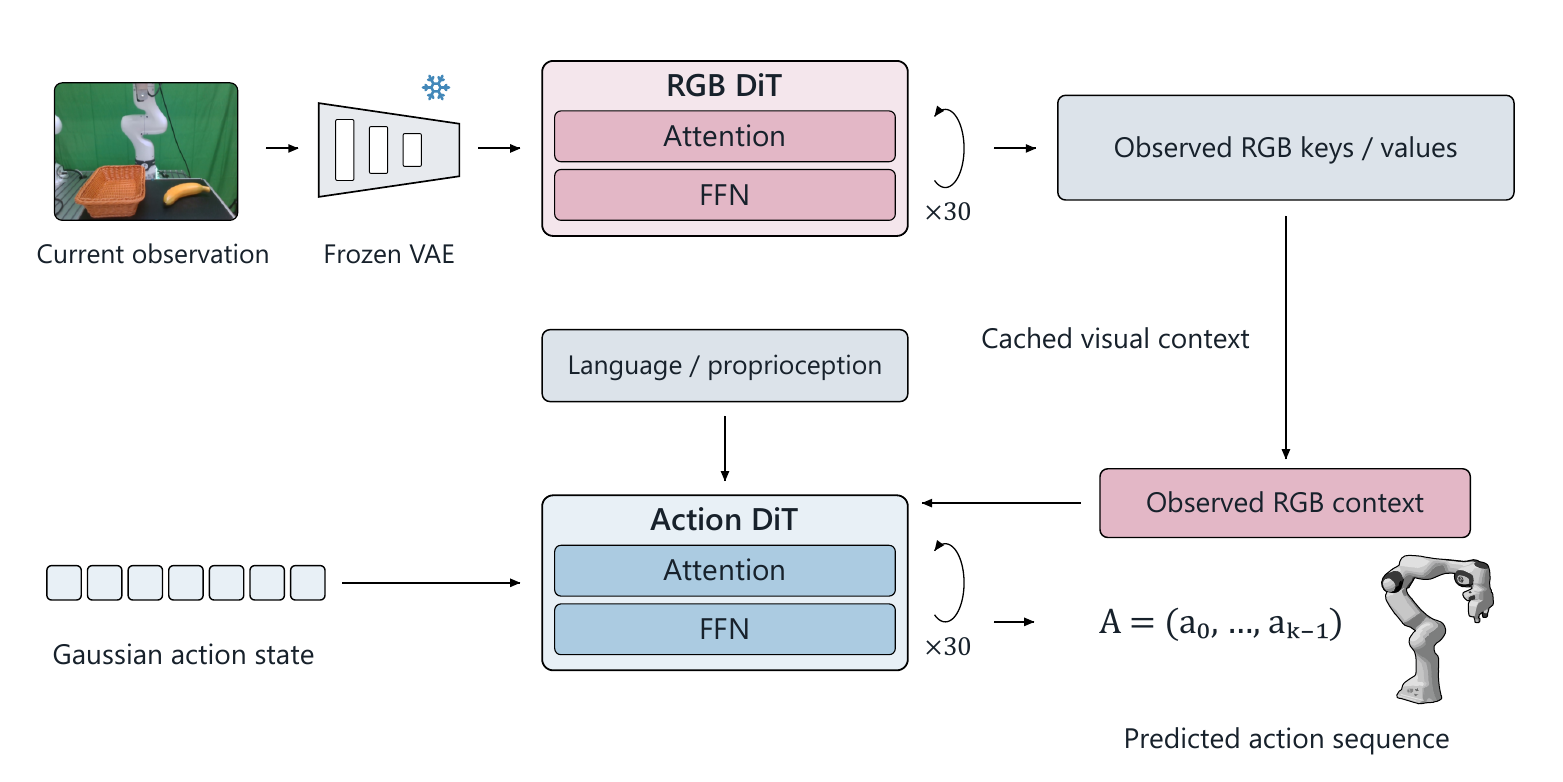}
    \captionof{figure}{HOG-WAM inference. The RGB DiT module encodes the current observation into cached keys and values. Conditioned on these features, language, and proprioception, the action DiT module integrates a velocity field from Gaussian noise to an action sequence $A$. Receding-horizon execution applies a prefix of the sequence before updating the observation. Auxiliary HOG prediction is confined to training, preserving the FastWAM inference architecture.}
    \label{fig:deployment}

\end{minipage}
\par\medskip

At inference, the HOG DiT and future visual tokens are omitted.
The current RGB canvas is encoded by the frozen VAE, and the observed RGB features are computed once with the instruction and proprioceptive context.
Their per-layer keys and values are cached while the action DiT integrates its velocity field from Gaussian noise to an action chunk.
With $N$ integration steps and a decreasing schedule $1=\sigma_N>\cdots>\sigma_0=0$, $A_i$ denotes the action chunk at noise level $\sigma_i$. A solver step is
\begin{equation}
    A_{i-1}=A_i+
    (\sigma_{i-1}-\sigma_i)
    \widehat{u}^a_\theta(A_i,\sigma_i\mid I_0,\ell,s_0).
\end{equation}
The output is transformed back to the action units used by the benchmark.
The controller executes the configured chunk prefix, receives a new observation, and replans.
For RoboTwin, FastWAM and HOG-WAM predict 32 absolute joint-position actions and execute 24 before replanning; for LIBERO, they predict 32 end-effector actions and execute eight.
Both policies use the same action representation and prediction and execution horizons within each benchmark.
HOG-WAM inference uses the RGB and action DiT modules.

\section{Real-robot tasks and evaluation}
\label{supp:real_robot}

\paragraph{Task suite.}
The four Franka tasks span reorientation, containment, stacking, and insertion.
Reorientation requires the white plastic bracket to remain in the intended upside-down pose; placement requires the target produce to lie fully inside the basket.
Stacking requires the red block to rest stably on the blue block; insertion requires the cylinder to be seated in its circular holder.
The main paper shows four recorded demonstration frames in the upper row and their corresponding gradient visualizations in the lower row, with one task per column.
These monochrome displays use the same Sobel-based visualization as the method figures; the three-channel HOG target is defined in \cref{supp:sec:hog_targets}.
\paragraph{Demonstrations and policy training.}
The collection contains 407,143 frames from 1,000 demonstrations recorded at 30 Hz.
Produce placement contributes 400 demonstrations, and each other task contributes 200.
Observations include images from a main camera and a wrist camera.
The state comprises seven measured joint positions and the gripper width; each action comprises seven absolute joint targets in radians and a binary gripper command.
Training preserves this sampling rate.
FastWAM and HOG-WAM use the same trajectory-level split of 900 training and 100 validation demonstrations, with normalization statistics computed only from training trajectories.
Both models use an effective batch size of 128, $\tau=8$, and a peak learning rate of $2.5\times10^{-5}$.
The learning rate uses a 300-step warmup and follows a 6,000-step cosine schedule through step 3,000.
It then decays continuously from that value to $2.5\times10^{-7}$ at step 30,000.
Optimizer state is preserved across the schedule transition.
The physical evaluation uses the final step-30,000 checkpoint for each model.

\paragraph{Observation conditions and scoring.}
The comparison uses RGB-input policies under clean observations and synthetic mixed noise.
The noisy condition applies the same mixed-noise operator as RoboTwin-Plus (\cref{supp:sec:evaluation}) to the composed RGB observation before visual encoding.
It combines additive Gaussian noise with standard deviation 10 in 8-bit intensity units and multiplicative speckle noise with standard deviation 0.05.
The resulting pixel values are rounded and clipped to $[0,255]$.
Language and proprioceptive conditioning are unchanged.
Each task--model--condition group contains 25 recorded binary outcomes, giving 100 trials per model and condition and 400 trials overall.
Each trial has a maximum budget of 200 steps and ends early when the task succeeds or a failure/error occurs.
After each trial, the robot is reset using the reset command.
All 25 recorded outcomes enter the denominator of their corresponding group.
\Cref{supp:real_robot_results} reports the success counts.
The macro-average assigns equal weight to the four tasks; because trial counts are equal, it also equals the pooled success rate within a model and condition.

\par\medskip
\noindent\begin{minipage}{\textwidth}
    \centering\small
    \setlength{\tabcolsep}{5pt}
    \captionof{table}{Completed Franka evaluation. Task entries give success counts out of 25 trials, followed by pointwise 95\% Wilson intervals for success rates (\%). Bold marks the higher count; ties remain unbolded. Totals report successes / 100 trials and the macro-average rate (\%). Intervals are unadjusted for multiple comparisons.}
    \label{supp:real_robot_results}
    \begin{checkedtable}{\textwidth}
    \begin{tabular}{@{}lcccc@{}}
        \toprule
        & \multicolumn{2}{c}{Clean} & \multicolumn{2}{c}{Noisy} \\
        \cmidrule(lr){2-3}\cmidrule(l){4-5}
        Task & FastWAM & HOG-WAM & FastWAM & HOG-WAM \\
        \midrule
        Bracket reorientation & 10 [23.4, 59.3] & \textbf{12} [30.0, 66.5] & 3 [4.2, 30.0] & \textbf{5} [8.9, 39.1] \\
        Produce placement & 15 [40.7, 76.6] & \textbf{17} [48.4, 82.8] & 8 [17.2, 51.6] & \textbf{10} [23.4, 59.3] \\
        Block stacking & 8 [17.2, 51.6] & \textbf{20} [60.9, 91.1] & 6 [11.5, 43.4] & \textbf{9} [20.2, 55.5] \\
        Cylinder insertion & 7 [14.3, 47.6] & \textbf{10} [23.4, 59.3] & 0 [0.0, 13.3] & 0 [0.0, 13.3] \\
        \midrule
        Total / macro-average & 40/100 (40) & \textbf{59/100 (59)} & 17/100 (17) & \textbf{24/100 (24)} \\
        \bottomrule
    \end{tabular}
    \end{checkedtable}
\end{minipage}
\par\medskip

\paragraph{Evaluation coverage.}
The comparison covers one trained checkpoint per model and one synthetic noise setting.
Trials are analyzed as unpaired observations because the records contain no matched reset identifiers.
The 30 Hz sampling rate describes the demonstrations; the evaluation records do not measure the executed action-update rate or document randomized model ordering.
\paragraph{Trial-level uncertainty.}
\Cref{supp:real_robot_results} gives pointwise 95\% Wilson score intervals for each task, model, and observation condition.
For $s$ successes in $n=25$ trials, let $\widehat p=s/n$ and $z=1.96$.
The interval endpoints are
\begin{equation}
\frac{\widehat p+z^2/(2n)\;\pm\;z\sqrt{\widehat p(1-\widehat p)/n+z^2/(4n^2)}}{1+z^2/n}.
\end{equation}
These intervals model independent binary trials within each task--model--condition group, conditional on the evaluated checkpoint and protocol.
They describe trial uncertainty; variability across independently trained policies is a separate quantity.
The zero-success insertion groups retain a positive upper bound, reflecting the finite trial count.

\FloatBarrier

\par\medskip
\noindent\begin{minipage}{\textwidth}
\section{Ablation study: complementary supervision and RGB replacement}
\label{supp:sec:ablation}
\subsection{Architectures and matched protocol}
\Cref{fig:replacement} contrasts auxiliary prediction with RGB replacement.

    \centering
    \includegraphics[width=.73\textwidth]{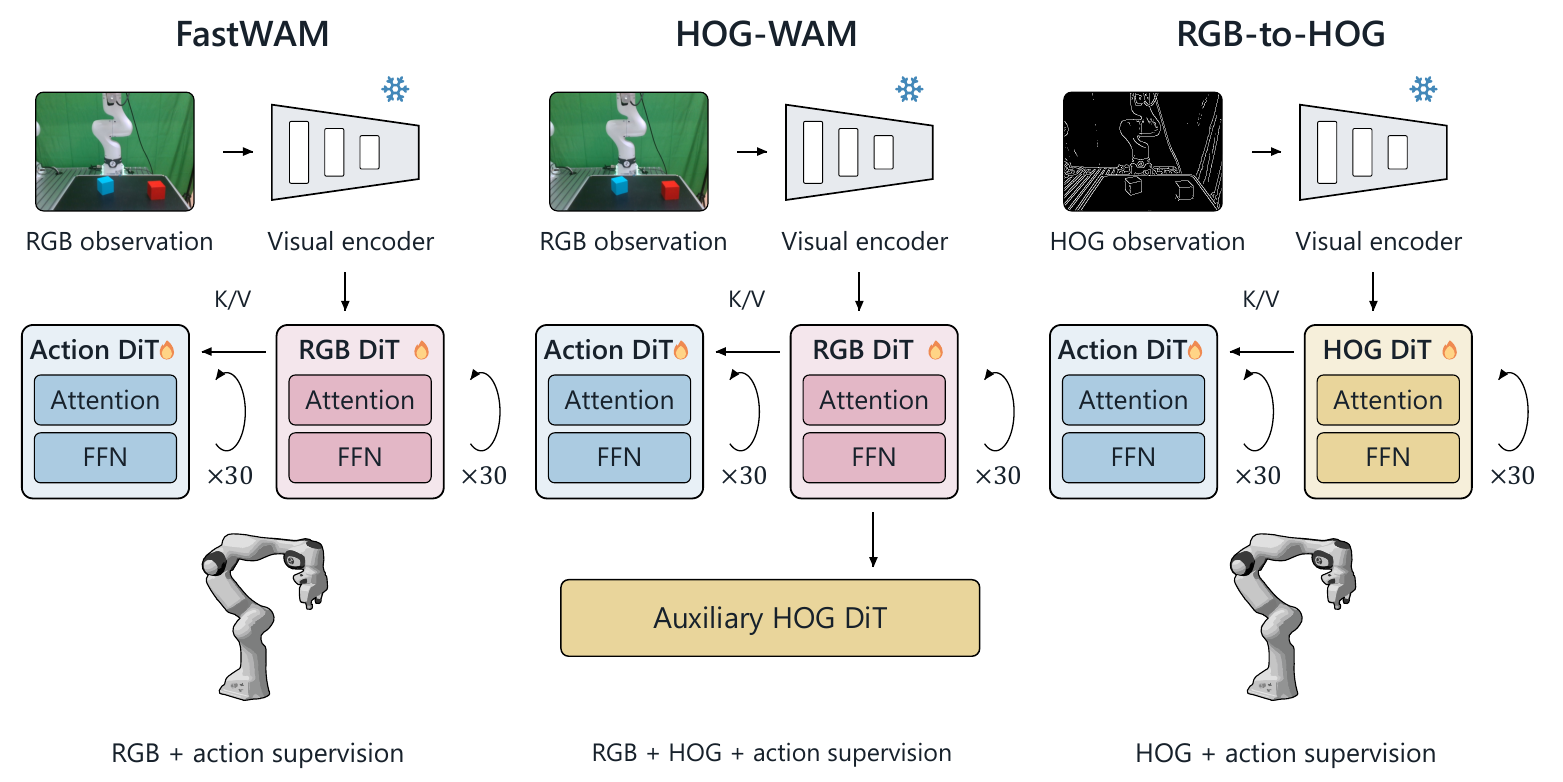}
    \captionof{figure}{Ablation of visual representation and predictive supervision. FastWAM jointly models future RGB and actions. HOG-WAM augments this formulation with an auxiliary HOG predictor that supervises the RGB backbone during training. RGB-to-HOG instead substitutes HOG for both visual observations and prediction targets within the visual and action DiT modules. HOG-WAM retains RGB-based inference, whereas the replacement model requires HOG observations at test time.}
    \label{fig:replacement}

\end{minipage}
\par\medskip

This study compares FastWAM, RGB-to-HOG, and HOG-WAM to distinguish auxiliary structural supervision from replacing RGB inputs and targets.
RGB-to-HOG retains the visual and action DiT modules of FastWAM and applies $\mathcal{H}$ before visual encoding during both training and inference.
Its visual DiT predicts future HOG latents in place of future RGB latents.
All three configurations use the same 40-task, 2,000-demonstration LIBERO collection and normalization statistics.
They share an effective batch size of 256, action-copy count $\tau=8$, peak learning rate of $10^{-4}$, weight decay of $10^{-2}$, and ten training epochs; all are evaluated at step 13,200.
FastWAM and RGB-to-HOG assign unit weights to their visual and action objectives; HOG-WAM additionally assigns unit weight to its HOG objective.
Each model uses the same 10,030 LIBERO Plus variants, instructions, first official initial states, random seeds, and action-step budgets.
Each policy predicts 32 actions with ten flow-integration steps and executes the first eight before replanning.

\subsection{Results by perturbation category}
\Cref{tab:libero_ablation_categories} reports the category breakdown.
RGB-to-HOG leads under background and lighting changes, while both RGB-input policies achieve higher overall success.

\par\medskip
\noindent\begin{minipage}{\textwidth}

\centering
\small
\setlength{\tabcolsep}{4pt}
\captionof{table}{Category-level representation ablation on LIBERO Plus. Cells report successes / success rate (\%). Category denominators, in row order, are 1,076, 1,599, 1,550, 1,537, 1,142, 1,525, and 1,601. The overall rate pools 10,030 trials. RGB replacement improves performance under texture and lighting shifts but reduces it in several other categories.}
\label{tab:libero_ablation_categories}
\begin{checkedtable}{\textwidth}
\begin{tabular}{@{}lrrr@{}}
\toprule
Category & FastWAM & RGB-to-HOG & HOG-WAM \\
\midrule
Background textures & 731 / 67.94 & \textbf{814 / 75.65} & 702 / 65.24 \\
Camera viewpoints & 1,033 / 64.60 & 691 / 43.21 & \textbf{1,098 / 68.67} \\
Robot initial states & \textbf{1,183 / 76.32} & 1,106 / 71.35 & 1,179 / 76.06 \\
Language instructions & \textbf{1,134 / 73.78} & 931 / 60.57 & 1,132 / 73.65 \\
Light conditions & 1,038 / 90.89 & \textbf{1,048 / 91.77} & 1,009 / 88.35 \\
Object layout & 1,165 / 76.39 & 1,085 / 71.15 & \textbf{1,166 / 76.46} \\
Sensor noise & 1,187 / 74.14 & 1,025 / 64.02 & \textbf{1,263 / 78.89} \\
\midrule
Overall & 7,471 / 74.49 & 6,700 / 66.80 & \textbf{7,549 / 75.26} \\
\bottomrule
\end{tabular}
\end{checkedtable}

\end{minipage}
\par\medskip

\clearpage
\section{Complete RoboTwin task results}
\Cref{tab:robotwin_counts,tab:robotwin_photometric_counts} report successes out of 100 episodes for each task, model, and condition.
P, M, F, and H denote $\pi_{0.5}$, MOTUS, FastWAM, and HOG-WAM, respectively.
The four corrupted clean-scene conditions constitute RoboTwin-Plus.

\noindent\begin{minipage}{\textwidth}

\centering
\footnotesize
\setlength{\tabcolsep}{3pt}
\renewcommand{\arraystretch}{0.96}
\captionof{table}{RoboTwin successes out of 100 episodes: native clean, native randomized, and clean with mixed noise. P: $\pi_{0.5}$; M: MOTUS; F: FastWAM; H: HOG-WAM. Noise combines Gaussian noise with a standard deviation of 10 in uint8 intensity levels and speckle noise with a standard deviation of 0.05. Bold indicates the highest count among the four models; ties are not bolded.}
\label{tab:robotwin_counts}
\begin{checkedtable}{\textwidth}
\begin{tabular}{@{}l*{12}{r}@{}}
\toprule
& \multicolumn{4}{c}{Clean} & \multicolumn{4}{c}{Randomized} & \multicolumn{4}{c}{Mixed noise} \\
\cmidrule(lr){2-5}\cmidrule(lr){6-9}\cmidrule(l){10-13}
Task & P & M & F & H & P & M & F & H & P & M & F & H \\
\midrule
\texttt{adjust\_bottle} & 92 & 97 & 100 & 100 & 97 & 85 & 100 & 100 & 86 & 63 & 100 & 100 \\
\texttt{beat\_block\_hammer} & 68 & 92 & 96 & \textbf{98} & 72 & 85 & 96 & 96 & 59 & 97 & 93 & 97 \\
\texttt{blocks\_ranking\_rgb} & 34 & 97 & \textbf{100} & 98 & 45 & 95 & \textbf{100} & 99 & 45 & 96 & 99 & \textbf{100} \\
\texttt{blocks\_ranking\_size} & 18 & 76 & 98 & 98 & 24 & 66 & \textbf{95} & 90 & 11 & 43 & 67 & \textbf{72} \\
\texttt{click\_alarmclock} & 75 & 100 & 100 & 100 & 69 & \textbf{100} & 99 & 99 & 63 & \textbf{100} & 99 & 92 \\
\texttt{click\_bell} & 58 & 100 & 100 & 100 & 59 & 99 & 100 & 100 & 46 & 100 & 98 & 100 \\
\texttt{dump\_bin\_bigbin} & 63 & 96 & 96 & \textbf{99} & 75 & 91 & 94 & \textbf{99} & 22 & \textbf{75} & 42 & 70 \\
\texttt{grab\_roller} & 85 & 100 & 100 & 100 & 86 & 99 & 99 & 99 & 92 & 96 & 100 & 100 \\
\texttt{handover\_block} & 4 & 77 & 89 & 89 & 11 & 67 & 85 & \textbf{89} & 3 & 24 & 12 & \textbf{67} \\
\texttt{handover\_mic} & 60 & 74 & 96 & \textbf{99} & 57 & 72 & 96 & \textbf{98} & 36 & 84 & \textbf{96} & 94 \\
\texttt{hanging\_mug} & 6 & 32 & 52 & 52 & 6 & 35 & \textbf{63} & 57 & 5 & 13 & 24 & \textbf{40} \\
\texttt{lift\_pot} & 27 & 99 & 100 & 100 & 18 & 98 & 99 & \textbf{100} & 8 & 94 & \textbf{99} & 97 \\
\texttt{move\_can\_pot} & 23 & 37 & \textbf{94} & 88 & 30 & 71 & 92 & \textbf{93} & 18 & 79 & 97 & \textbf{100} \\
\texttt{move\_pillbottle\_pad} & 32 & 99 & \textbf{100} & 99 & 41 & 98 & \textbf{100} & 99 & 27 & \textbf{86} & 69 & 74 \\
\texttt{move\_playingcard\_away} & 79 & 98 & 100 & 100 & 81 & 97 & 99 & \textbf{100} & 66 & 100 & 98 & 100 \\
\texttt{move\_stapler\_pad} & 19 & 76 & \textbf{85} & 82 & 24 & \textbf{79} & 74 & 68 & 9 & \textbf{67} & 52 & 45 \\
\texttt{open\_laptop} & 62 & 95 & 96 & \textbf{99} & 67 & 87 & \textbf{100} & 99 & 53 & 73 & 89 & \textbf{97} \\
\texttt{open\_microwave} & 56 & \textbf{94} & 66 & 49 & 51 & \textbf{94} & 69 & 60 & 16 & \textbf{89} & 22 & 36 \\
\texttt{pick\_diverse\_bottles} & 32 & \textbf{89} & 85 & 85 & 27 & \textbf{88} & 81 & 80 & 16 & 73 & \textbf{89} & 84 \\
\texttt{pick\_dual\_bottles} & 34 & 90 & 90 & \textbf{94} & 30 & 92 & 88 & \textbf{93} & 24 & 84 & 97 & 97 \\
\texttt{place\_a2b\_left} & 57 & 91 & \textbf{99} & 97 & 64 & 82 & 95 & 95 & 45 & 69 & 81 & \textbf{93} \\
\texttt{place\_a2b\_right} & 59 & 86 & \textbf{99} & 97 & 70 & 78 & 97 & 97 & 50 & 75 & 86 & \textbf{88} \\
\texttt{place\_bread\_basket} & 29 & \textbf{96} & 93 & 92 & 60 & \textbf{97} & 95 & 89 & 32 & 85 & 84 & \textbf{91} \\
\texttt{place\_bread\_skillet} & 28 & 91 & 94 & \textbf{96} & 22 & 87 & 90 & \textbf{92} & 16 & \textbf{90} & 87 & 81 \\
\texttt{place\_burger\_fries} & 62 & 97 & 99 & 99 & 53 & 94 & \textbf{99} & 94 & 36 & 93 & \textbf{96} & 92 \\
\texttt{place\_can\_basket} & 24 & \textbf{78} & 58 & 63 & 33 & \textbf{69} & 62 & 67 & 19 & \textbf{81} & 59 & 47 \\
\texttt{place\_cans\_plasticbox} & 36 & 92 & \textbf{97} & 96 & 40 & 95 & \textbf{100} & 99 & 27 & 96 & \textbf{99} & 98 \\
\texttt{place\_container\_plate} & 82 & 97 & \textbf{100} & 99 & 85 & \textbf{100} & 99 & 98 & 77 & 97 & 97 & \textbf{98} \\
\texttt{place\_dual\_shoes} & 21 & \textbf{95} & 89 & 85 & 28 & 88 & \textbf{89} & 85 & 2 & 63 & 67 & \textbf{77} \\
\texttt{place\_empty\_cup} & 63 & 96 & 100 & 100 & 61 & 98 & 99 & \textbf{100} & 66 & 95 & 90 & 95 \\
\texttt{place\_fan} & 41 & 97 & 93 & \textbf{98} & 52 & 86 & 91 & 91 & 16 & 46 & \textbf{69} & 64 \\
\texttt{place\_mouse\_pad} & 15 & 71 & \textbf{96} & 90 & 20 & 62 & \textbf{89} & 85 & 11 & 60 & 58 & \textbf{64} \\
\texttt{place\_object\_basket} & 36 & 86 & \textbf{89} & 87 & 43 & 82 & \textbf{86} & 79 & 28 & 79 & \textbf{92} & 87 \\
\texttt{place\_object\_scale} & 41 & 87 & 94 & \textbf{98} & 46 & 81 & 92 & \textbf{95} & 34 & 71 & 80 & \textbf{89} \\
\texttt{place\_object\_stand} & 69 & 96 & 96 & 95 & 77 & \textbf{100} & 88 & 91 & 63 & \textbf{98} & 89 & 95 \\
\texttt{place\_phone\_stand} & 40 & 91 & 94 & \textbf{96} & 42 & 85 & 96 & \textbf{97} & 40 & 84 & 86 & 86 \\
\texttt{place\_shoe} & 54 & \textbf{99} & 93 & 94 & 55 & \textbf{97} & 95 & 96 & 34 & \textbf{91} & 80 & 72 \\
\texttt{press\_stapler} & 57 & \textbf{94} & 89 & 85 & 55 & \textbf{97} & 92 & 88 & 50 & \textbf{94} & 88 & 89 \\
\texttt{put\_bottles\_dustbin} & 16 & 88 & 89 & \textbf{91} & 17 & 84 & 85 & \textbf{90} & 4 & 47 & 15 & \textbf{69} \\
\texttt{put\_object\_cabinet} & 28 & 56 & 90 & \textbf{94} & 37 & 52 & 86 & \textbf{95} & 25 & 28 & \textbf{64} & 62 \\
\texttt{rotate\_qrcode} & 37 & 82 & 88 & \textbf{94} & 51 & 75 & \textbf{90} & 87 & 24 & 74 & 74 & \textbf{77} \\
\texttt{scan\_object} & 19 & 66 & 92 & 92 & 28 & 62 & 86 & \textbf{89} & 16 & 67 & 78 & \textbf{81} \\
\texttt{shake\_bottle} & 98 & 100 & 100 & 100 & 97 & 96 & \textbf{100} & 99 & 94 & 94 & 99 & \textbf{100} \\
\texttt{shake\_bottle\_horizontally} & 97 & 100 & 100 & 100 & 96 & 97 & 100 & 100 & 94 & 97 & 100 & 100 \\
\texttt{stack\_blocks\_three} & 27 & 85 & \textbf{95} & 92 & 31 & 92 & 93 & \textbf{95} & 15 & 87 & 90 & \textbf{93} \\
\texttt{stack\_blocks\_two} & 57 & 100 & 100 & 100 & 57 & 96 & \textbf{100} & 99 & 54 & 99 & 100 & 100 \\
\texttt{stack\_bowls\_three} & 41 & 80 & \textbf{91} & 80 & 36 & \textbf{88} & 81 & 83 & 32 & 75 & \textbf{76} & 72 \\
\texttt{stack\_bowls\_two} & 80 & \textbf{97} & 95 & 96 & 74 & 97 & 97 & \textbf{98} & 79 & 93 & \textbf{96} & 93 \\
\texttt{stamp\_seal} & 33 & \textbf{93} & 87 & 91 & 39 & 93 & 93 & \textbf{97} & 19 & \textbf{81} & 75 & 73 \\
\texttt{turn\_switch} & 25 & \textbf{78} & 62 & 68 & 30 & 80 & 71 & 80 & 23 & 53 & 46 & \textbf{55} \\
\midrule
Total / 5,000 & 2299 & 4383 & 4604 & 4594 & 2469 & 4288 & 4565 & 4568 & 1830 & 3898 & 3943 & 4143 \\
\bottomrule
\end{tabular}
\end{checkedtable}

\end{minipage}\par
\clearpage
\noindent\begin{minipage}{\textwidth}

\centering
\footnotesize
\setlength{\tabcolsep}{3pt}
\renewcommand{\arraystretch}{0.96}
\captionof{table}{RoboTwin successes out of 100 episodes: clean scenes with warm filtering, underexposure, and contrast reduction. P: $\pi_{0.5}$; M: MOTUS; F: FastWAM; H: HOG-WAM. Together with mixed noise, these conditions form RoboTwin-Plus. Bold indicates the highest count among the four models; ties are not bolded.}
\label{tab:robotwin_photometric_counts}
\begin{checkedtable}{\textwidth}
\begin{tabular}{@{}l*{12}{r}@{}}
\toprule
& \multicolumn{4}{c}{Warm filter} & \multicolumn{4}{c}{Underexposure} & \multicolumn{4}{c}{Reduced contrast} \\
\cmidrule(lr){2-5}\cmidrule(lr){6-9}\cmidrule(l){10-13}
Task & P & M & F & H & P & M & F & H & P & M & F & H \\
\midrule
\texttt{adjust\_bottle} & 94 & 96 & 100 & 100 & 95 & 99 & 100 & 100 & 95 & 92 & 100 & 100 \\
\texttt{beat\_block\_hammer} & 72 & 93 & \textbf{98} & 97 & 69 & 91 & 98 & \textbf{99} & 76 & 96 & \textbf{98} & 97 \\
\texttt{blocks\_ranking\_rgb} & 33 & \textbf{100} & 99 & 99 & 38 & 97 & 99 & \textbf{100} & 45 & 98 & 100 & 100 \\
\texttt{blocks\_ranking\_size} & 18 & 76 & 88 & \textbf{90} & 28 & 73 & \textbf{95} & 93 & 14 & 74 & 90 & 90 \\
\texttt{click\_alarmclock} & 74 & 100 & 100 & 100 & 68 & 100 & 100 & 99 & 75 & 100 & 100 & 100 \\
\texttt{click\_bell} & 60 & 100 & 100 & 100 & 64 & 100 & 100 & 100 & 61 & 100 & 100 & 100 \\
\texttt{dump\_bin\_bigbin} & 65 & \textbf{100} & 91 & 92 & 57 & 95 & 96 & 96 & 62 & 93 & 78 & 93 \\
\texttt{grab\_roller} & 86 & 100 & 100 & 100 & 89 & 100 & 100 & 100 & 84 & 100 & 100 & 100 \\
\texttt{handover\_block} & 5 & 79 & \textbf{93} & 90 & 10 & 79 & \textbf{93} & 88 & 8 & 81 & 85 & \textbf{92} \\
\texttt{handover\_mic} & 49 & 74 & 98 & \textbf{100} & 59 & 67 & 97 & 97 & 56 & 76 & 98 & \textbf{100} \\
\texttt{hanging\_mug} & 7 & 30 & 41 & \textbf{56} & 4 & 26 & 48 & \textbf{61} & 6 & 23 & 50 & \textbf{51} \\
\texttt{lift\_pot} & 21 & 98 & 100 & 100 & 27 & 100 & 100 & 99 & 28 & 99 & \textbf{100} & 99 \\
\texttt{move\_can\_pot} & 26 & 22 & 84 & \textbf{89} & 27 & 43 & \textbf{96} & 86 & 28 & 36 & 88 & \textbf{91} \\
\texttt{move\_pillbottle\_pad} & 32 & 98 & 97 & \textbf{100} & 36 & 98 & \textbf{100} & 99 & 31 & 97 & 99 & \textbf{100} \\
\texttt{move\_playingcard\_away} & 74 & 98 & 99 & 99 & 83 & 99 & 100 & 100 & 81 & 100 & 100 & 98 \\
\texttt{move\_stapler\_pad} & 17 & \textbf{76} & 69 & 70 & 21 & 80 & \textbf{83} & 82 & 15 & \textbf{79} & 74 & 74 \\
\texttt{open\_laptop} & 66 & 94 & 99 & 99 & 63 & 94 & 98 & 98 & 61 & 95 & 99 & 99 \\
\texttt{open\_microwave} & 51 & \textbf{90} & 65 & 28 & 53 & \textbf{95} & 75 & 64 & 55 & \textbf{96} & 47 & 28 \\
\texttt{pick\_diverse\_bottles} & 29 & \textbf{87} & 76 & 84 & 35 & \textbf{88} & 78 & 86 & 34 & 93 & 80 & \textbf{94} \\
\texttt{pick\_dual\_bottles} & 32 & 92 & 80 & \textbf{93} & 32 & 89 & 79 & \textbf{90} & 34 & 94 & 80 & \textbf{95} \\
\texttt{place\_a2b\_left} & 52 & 76 & 88 & \textbf{92} & 56 & 86 & 98 & 98 & 57 & \textbf{90} & 86 & 89 \\
\texttt{place\_a2b\_right} & 59 & 73 & 96 & 96 & 62 & 83 & 94 & \textbf{96} & 54 & 75 & \textbf{95} & 92 \\
\texttt{place\_bread\_basket} & 32 & 92 & \textbf{95} & 93 & 34 & 91 & \textbf{96} & 93 & 33 & 92 & \textbf{98} & 96 \\
\texttt{place\_bread\_skillet} & 25 & 88 & \textbf{92} & 90 & 34 & 88 & 87 & \textbf{93} & 21 & 89 & \textbf{100} & 91 \\
\texttt{place\_burger\_fries} & 55 & 96 & 99 & 99 & 66 & 96 & 98 & 98 & 67 & 98 & 99 & 99 \\
\texttt{place\_can\_basket} & 23 & \textbf{81} & 50 & 57 & 25 & \textbf{81} & 45 & 58 & 35 & \textbf{80} & 45 & 45 \\
\texttt{place\_cans\_plasticbox} & 25 & 92 & 94 & \textbf{99} & 17 & 97 & 97 & 97 & 25 & 95 & 96 & \textbf{100} \\
\texttt{place\_container\_plate} & 83 & 98 & 95 & \textbf{100} & 80 & \textbf{99} & 98 & 98 & 83 & 100 & 100 & 100 \\
\texttt{place\_dual\_shoes} & 22 & \textbf{93} & 88 & 80 & 21 & \textbf{91} & 84 & 79 & 24 & \textbf{92} & 89 & 87 \\
\texttt{place\_empty\_cup} & 59 & 99 & 100 & 100 & 70 & 98 & 100 & 100 & 59 & 99 & 99 & \textbf{100} \\
\texttt{place\_fan} & 39 & 93 & 93 & \textbf{94} & 54 & \textbf{96} & 93 & 94 & 50 & 89 & 86 & \textbf{93} \\
\texttt{place\_mouse\_pad} & 13 & 65 & \textbf{93} & 91 & 16 & 70 & \textbf{93} & 89 & 19 & 67 & \textbf{90} & 88 \\
\texttt{place\_object\_basket} & 39 & 80 & \textbf{88} & 84 & 45 & 80 & 84 & 84 & 40 & 85 & 87 & \textbf{88} \\
\texttt{place\_object\_scale} & 37 & 88 & 91 & \textbf{93} & 32 & 85 & 95 & \textbf{98} & 40 & 92 & 89 & \textbf{95} \\
\texttt{place\_object\_stand} & 72 & \textbf{97} & 91 & 93 & 75 & \textbf{97} & 92 & 96 & 70 & \textbf{98} & 94 & 96 \\
\texttt{place\_phone\_stand} & 46 & 90 & 91 & \textbf{93} & 53 & 91 & 94 & \textbf{95} & 51 & 91 & 89 & 91 \\
\texttt{place\_shoe} & 52 & \textbf{99} & 93 & 87 & 51 & \textbf{99} & 93 & 92 & 54 & \textbf{99} & 91 & 88 \\
\texttt{press\_stapler} & 59 & \textbf{93} & 87 & 92 & 60 & \textbf{93} & 89 & 84 & 63 & \textbf{95} & 87 & 83 \\
\texttt{put\_bottles\_dustbin} & 13 & 88 & 86 & \textbf{92} & 29 & 86 & 90 & \textbf{94} & 19 & 88 & 86 & \textbf{94} \\
\texttt{put\_object\_cabinet} & 24 & 54 & 79 & \textbf{85} & 28 & 61 & 93 & \textbf{96} & 32 & 59 & 84 & \textbf{86} \\
\texttt{rotate\_qrcode} & 30 & 84 & 82 & \textbf{88} & 37 & 82 & 90 & \textbf{91} & 40 & 77 & \textbf{90} & 89 \\
\texttt{scan\_object} & 13 & 68 & 90 & \textbf{93} & 17 & 80 & \textbf{91} & 90 & 16 & 69 & \textbf{89} & 88 \\
\texttt{shake\_bottle} & 95 & 100 & 100 & 100 & 99 & 100 & 100 & 100 & 97 & 100 & 100 & 100 \\
\texttt{shake\_bottle\_horizontally} & 97 & 100 & 100 & 100 & 97 & 100 & 99 & 100 & 96 & 100 & 100 & 100 \\
\texttt{stack\_blocks\_three} & 26 & 78 & \textbf{99} & 94 & 27 & 84 & 93 & \textbf{94} & 21 & 87 & \textbf{95} & 94 \\
\texttt{stack\_blocks\_two} & 50 & 100 & 100 & 99 & 61 & \textbf{100} & 99 & 98 & 61 & 100 & 100 & 99 \\
\texttt{stack\_bowls\_three} & 35 & 84 & 82 & \textbf{85} & 49 & 73 & 74 & \textbf{80} & 39 & 79 & 82 & \textbf{84} \\
\texttt{stack\_bowls\_two} & 79 & 96 & 96 & \textbf{99} & 75 & 96 & \textbf{98} & 96 & 75 & \textbf{99} & 95 & 94 \\
\texttt{stamp\_seal} & 39 & \textbf{91} & 80 & 79 & 40 & \textbf{95} & 90 & 91 & 35 & \textbf{93} & 80 & 79 \\
\texttt{turn\_switch} & 27 & \textbf{75} & 63 & 65 & 34 & \textbf{72} & 63 & 65 & 25 & \textbf{77} & 70 & 71 \\
\midrule
Total / 5,000 & 2231 & 4314 & 4458 & 4498 & 2402 & 4363 & 4545 & 4574 & 2350 & 4376 & 4457 & 4500 \\
\bottomrule
\end{tabular}
\end{checkedtable}

\end{minipage}\par

\FloatBarrier
\clearpage

\end{document}